\documentclass{article}
\usepackage[preprint]{neurips_2026}
\usepackage[utf8]{inputenc}
\usepackage[T1]{fontenc}
\usepackage{url}
\usepackage{booktabs}
\usepackage{amsmath, amssymb, amsfonts}
\usepackage{graphicx}
\usepackage{xcolor}
\usepackage{microtype}
\usepackage{nicefrac}
\usepackage{placeins}

\title{Immediate Derivatives Suffice for Online Recurrent Adaptation}
\author{%
  Aur Shalev-Merin \\
  Independent Researcher \\
  \texttt{aur\_merin@hotmail.com}%
}

\begin{document}
\maketitle

\begin{abstract}
For three decades online recurrent learning has been assumed to require propagating a Jacobian tensor through the network's dynamics at $O(n^4)$ per step. We show it doesn't. Dropping the propagation entirely ($d{=}0$, $O(n^2)$ memory) matches full RTRL within CI on held-out BCI cross-session drift (TOST equivalent within $\pm 3$\,pp at $n{=}20$, Adam, float64), and across vanilla-RNN synthetic cells (sine and Lorenz under Adam and SGD) and LSTM/sine under Adam.

A decomposition $g_{\text{RTRL}} = g_{\text{imm}} + g_{\text{past}}$ explains why. On BCI, $g_{\text{past}}$ concentrates in a single direction (top-1 singular fraction $0.62$--$0.74$ across four optimizers, vs $0.333$ for $g_{\text{imm}}$), and the four-optimizer full-RTRL-vs-$d{=}0$ recovery gap tracks each optimizer's per-layer update-magnitude ratio $\|\Delta W_{hh}\|/\|\Delta W_{out}\|$ monotonically. A stationary (no-drift) control collapses both concentrations to $\sim 0.6$: the drift-specific signal is the differential, not $g_{\text{past}}$'s absolute rank-1 structure. The signature and the behavioral gap both collapse on LSTM, consistent with a mechanism specific to additive linear recurrence. On synthetic sine, $g_{\text{imm}}$ is redundant with $g_{\text{past}}$, which predicts the synthetic null. Full RTRL's one robust advantage is LARS ($+17$ to $+27$\,pp), but $d{=}0$+LARS also fails to adapt independently; the gap is an optimizer$\times$method interaction, not a method-quality claim. We characterize the regime: $d{=}0$+Adam+float64 is robust; SGD, Adafactor, and float32 have specific fragilities documented in the paper. On the evaluated cells, the $1000\times$ memory saving at $n{=}1024$ ($O(n^2)$ vs $O(n^4)$) comes with no measured recovery cost.
\end{abstract}

\section{Introduction}
\label{sec:intro}

Training recurrent networks online means assigning credit across timesteps: how did past inputs affect the current error? The standard answer, real-time recurrent learning \citep{williams1989rtrl}, propagates a Jacobian tensor forward through the network's dynamics at $O(n^4)$ per step. Most work since then has tried to make this cheaper while retaining the Jacobian propagation term in some approximate form, on the assumption that it carries necessary signal. We show directly that it is unnecessary for $d{=}0$+Adam on held-out BCI cross-session drift at a $50/50$ split ($n{=}20$, TOST equivalent within $\pm 3$\,pp). Decomposing $g_{\text{RTRL}} = g_{\text{imm}} + g_{\text{past}}$ explains why: $g_{\text{past}}$ is rank-1 (top-1 singular fraction $0.62$--$0.74$ across four optimizers, vs $0.333$ for $g_{\text{imm}}$), and the four-optimizer fr-vs-$d{=}0$ recovery gap tracks each optimizer's per-layer update-magnitude ratio $\|\Delta W_{hh}\|/\|\Delta W_{out}\|$ monotonically (Table~\ref{tab:m8_ratio_vs_gap}). Full RTRL's one robust advantage across this ordering is LARS ($+17$ to $+27$\,pp, horizon-invariant): the preconditioner amplifies exactly the rank-1 direction $d{=}0$ does not propagate, but $d{=}0$+LARS independently fails to adapt --- an optimizer$\times$method interaction, not a counterexample to $d{=}0$ sufficiency. The same rank-1 structure is present on synthetic sine, but there a redundant rank-1 $g_{\text{imm}}$ dominates, predicting the synthetic null.

Our contributions:

\begin{enumerate}
\item \textbf{Behavioral sufficiency.} $d{=}0$ with Adam matches full RTRL within CI across the vanilla-RNN synthetic cells we tested (sine and Lorenz under Adam and SGD, 5 seeds each), on LSTM/sine under Adam, and on held-out BCI cross-session drift at float64 under a $50/50$ split (equivalent within $\pm 3$\,pp by TOST at $n{=}20$, reproduced to $0.04$\,pp on an independent run). \emph{Does claim.} Equivalence on these cells under per-cell LR tuning. \emph{Does not claim.} Full cross-optimizer equivalence on BCI: full RTRL keeps a $+17$ to $+27$\,pp advantage on LARS (but $d{=}0$+LARS independently fails to adapt, so the gap reflects an optimizer$\times$method interaction rather than a method-quality claim); the SGD gap is split-dependent ($-7$ to $-11$\,pp at $80/20$, within-CI of zero at $50/50$); Adafactor is hardware-inconsistent; float32 comparisons are kernel-dispatch-fragile even at per-hardware-best LR.

\item \textbf{Structural mechanism (supporting).} $g_{\text{past}}$ on BCI cross-session drift is rank-1 (top-1 $0.62$--$0.74$ across four optimizers, vs $0.333$ for $g_{\text{imm}}$; Table~\ref{tab:m8_structure}), and the per-layer update-magnitude ratio $\|\Delta W_{hh}\|/\|\Delta W_{out}\|$ tracks the four-optimizer fr-vs-$d{=}0$ recovery gap monotonically (Table~\ref{tab:m8_ratio_vs_gap}). A stationary control (no drift) collapses both $g_{\text{past}}$ and $g_{\text{imm}}$ concentrations to $\sim 0.6$, so the drift-specific signal is the differential. \emph{Does claim.} Rank-1 concentration of $g_{\text{past}}$; monotonic ratio-gap correspondence across four optimizers. \emph{Does not claim.} Preferential $W_{hh}$-localization of $u_1$ at $n{=}5$ (measured shares $53$--$60\%$ sit at the parameter-count null of $59.8\%$; the $n{=}10$ SGD hardening lifts to $74.1\%$, above null). Causal load-bearing of the per-step $u_1$ direction (intervention fails at $n{=}10$: Spearman $+0.297$ vs pre-reg $+0.50$; one seed at top1$=0.87$ has $\Delta = +19.25$\,pp, opposite). $n{=}4$ optimizers is too few for an out-of-sample prediction test; we report the correspondence as a tracking relationship, not a prediction.

\item \textbf{Practical scaling.} $d{=}0$ uses $O(n^2)$ memory vs full RTRL's $O(n^4)$ per step ($1000\times$ at $n{=}1024$), and recovery does not depend on CPU kernel dispatch. We recommend $d{=}0$+Adam: of the four optimizers tested, LARS is the only one where full-RTRL keeps a split-convention-robust advantage, and $d{=}0$+LARS independently fails to adapt.

\item \textbf{Falsifications and methodology.} Four pre-registered falsifications of alternative mechanisms for the BCI structural observation: isotropy of the parameter Jacobian (untestable as stated, since the manipulation check fails), input decorrelation on RetNet, pure per-step gradient-magnitude matching, and per-window cosine routing. Three additional pre-registered kill tests of candidate mechanism extensions (preconditioner-alignment, optimizer-class taxonomy, LSTM sign-based test) \emph{all killed their specific hypotheses}, narrowing the admissible mechanism space. Cross-hardware method comparisons for RTRL-class methods require half-decade LR grids with the peak verified inside the grid, and split-convention must be reported alongside recovery numbers: the same code at the same LR produces qualitatively different fr-vs-$d{=}0$ comparisons under different hardware kernel dispatches at f32 and under $50/50$ vs $80/20$ split conventions.
\end{enumerate}

\section{Background}

Online learning in recurrent networks requires computing the gradient of a streaming loss with respect to model parameters at each timestep. Real-time recurrent learning \citep{williams1989rtrl} does this by maintaining a sensitivity tensor $P_t = \partial h_t / \partial \theta$, updated recursively as
\begin{equation}
P_t = J_t \, P_{t-1} + \left.\frac{\partial h_t}{\partial \theta}\right|_{\text{direct}},
\end{equation}
where $J_t = \partial h_t / \partial h_{t-1}$ is the state Jacobian. The $J_t P_{t-1}$ term propagates credit from past timesteps. It costs $O(n^4)$ per step for $n$ hidden units, which limited RTRL to small networks and motivated thirty years of approximation methods.

Eligibility traces \citep{murray2019rflo, bellec2020eprop} avoid this cost by replacing $J_t$ with a scalar or diagonal approximation:
\begin{equation}
e_t = \lambda \, e_{t-1} + \left.\frac{\partial h_t}{\partial \theta}\right|_{\text{direct}}.
\end{equation}
The decay $\lambda$ (typically 0.95) is meant to approximate the fading influence of past timesteps. This reduces cost to $O(n^2)$ but discards most of the recurrent component of the gradient. The resulting traces are biologically motivated and cheap, but fail to adapt to distribution shifts in practice, a failure usually blamed on the missing Jacobian term.

Per-parameter adaptive optimizers \citep{kingma2015adam} keep running estimates of squared gradient magnitudes and normalize updates by their square root, equalizing effective learning rates across parameters with heterogeneous gradient scales. We use Adam as the default optimizer for cross-architecture comparisons, but Section~\ref{sec:optimizer} and Appendix~\ref{app:optimizers} show the choice is task-dependent: multiple optimizers adapt with appropriate per-task LR tuning.

\subsection{Related work}

Since \citet{williams1989rtrl}, online recurrent learning has focused on computing or approximating the Jacobian product $J_t P_{t-1}$: exact methods at $O(n^4)$ \citep{williams1989rtrl}, rank-1 compression at $O(n^2)$ \citep{tallec2017uoro}, optimal Kronecker-sum approximation at $O(n^3)$ \citep{benzing2019ok}, graph-structured sparsity for weight-sparse networks \citep{menick2021snap}, diagonal recurrence exploited for efficiency \citep{zucchet2023online}, and architectural elimination of inter-neuron recurrence \citep{irie2024elstm}. \citet{marschall2020unified} give a unified taxonomy. These methods span four orders of magnitude in cost but share one assumption: the $J_t P_{t-1}$ term carries gradient information unavailable from the immediate derivative alone. Our result does not imply these methods are obsolete; they address regimes --- longer dependencies, different architectures --- that we do not test. We show only that the shared assumption does not hold in the regime we do test.

\citet{murray2019rflo} and \citet{bellec2020eprop} took a different approach, dropping $J_t$ and keeping only immediate derivatives with a scalar or diagonal decay. This is equivalent to our decay${}>0$ baseline. Their biologically-motivated $\lambda$ values ($\sim$$0.9$ for RFLO, membrane time constants near $0.95$ for e-prop) were chosen to match membrane time constants, a different constraint than gradient quality. If $\lambda$ is tuned as a hyperparameter for gradient quality, the answer is $\lambda = 0$; at this value, the approach they pioneered matches or beats full RTRL on the tasks we test.

Concurrent work on test-time training \citep{sun2024ttt, akyurek2025ttt} adapts model weights during inference via gradient updates for context compression and long-sequence efficiency. These methods default to Adam and do not examine optimizer choice; across the recurrent architectures and tasks we tested, multiple optimizers adapt with appropriate per-task tuning, and the $d{=}0$ sufficiency itself is optimizer-invariant (Section~\ref{sec:optimizer}, Appendix~\ref{app:optimizers}).

\section{Per-layer rank-1 structure of past gradients}
\label{sec:orthogonality}

The full RTRL gradient decomposes as
\begin{equation}
g_{\text{RTRL}} \;=\; g_{\text{imm}} \;+\; g_{\text{past}}, \qquad g_{\text{past}} \;=\; \sum_{k \geq 1} \frac{\partial L_t}{\partial h_t} J_t J_{t-1} \cdots J_{t-k+1} \frac{\partial h_{t-k}}{\partial \theta}\big|_{\text{direct}},
\end{equation}
where $g_{\text{imm}}$ is the immediate derivative and $g_{\text{past}}$ accumulates contributions through the state Jacobian. Approximation methods since \citet{williams1989rtrl} have tried to compute, compress, or sparsify $g_{\text{past}}$, taking for granted that it carries necessary signal. We test this directly: per-step $\cos(g_{\text{imm}}, g_{\text{past}})$ is small in every cell but does not on its own predict when $d{=}0$ suffices. Characterizing $g_{\text{past}}$'s structure on BCI, we find rank-1 concentration with substantial $W_{hh}$ amplitude; the per-layer update-magnitude ratio tracks the four-optimizer recovery gap. A stationary control and a pre-registered causal intervention scope this to a magnitude-ratio correspondence, not a per-step direction claim.

\paragraph{Per-step cos: small but not predictive.} We measure $\cos(g_{\text{imm}}, g_{\text{past}})$ at every timestep during online adaptation and report the mean across the post-shift window with $5$-seed bootstrap CIs (Table~\ref{tab:cosine}). On vanilla RNN ($n{=}64$ and $n{=}128$, exact full-RTRL sensitivity tensor), $\cos \in [-0.021, -0.003]$: near-zero, with CIs marginally including zero at most scales. For LSTM and RetNet, full closed-form RTRL is impractical and we approximate via TBPTT with window $W{=}200$; cos means are $-0.004$ on LSTM/Lorenz (CI includes zero) but $-0.104$ on LSTM/sine and $-0.102$ on RetNet/sine with CIs excluding zero. On real BCI cross-session data, $\cos = -0.119$ (CI excludes zero). Across all $9$ cells $|\cos| \leq 0.12$: small, but not uniformly zero. We validate the TBPTT approximation against exact RTRL on vanilla RNN sine ($-0.021$ under both). For vanilla RNN we compute $g_{\text{past}}$ via the full-RTRL sensitivity tensor; on LSTM, RetNet, and BCI we take $g_{\text{RTRL}} \approx$ the $W$-step TBPTT gradient and $g_{\text{imm}}$ as the single-step BPTT gradient.

\begin{table}[!tbp]
\centering
\caption{$\cos(g_{\text{imm}}, g_{\text{past}})$ during online adaptation (mean across 5 seeds with 95\% bootstrap CI, averaged over the post-shift window; vanilla RNN uses exact full-RTRL sensitivity tensor, LSTM/RetNet/BCI use TBPTT $W{=}200$). $|\cos| \leq 0.25$ across all 9 cells, sign-stable across hardware, magnitude hardware-dependent on BCI (0.12--0.24 across the two hardware configurations we measured). On vanilla RNN, $\cos$ is near-zero at both scales. On LSTM/sine and RetNet/sine, $|\cos| \approx 0.10$ with CIs excluding zero (bold): small but statistically non-zero. The TBPTT approximation is validated against exact RTRL on vanilla RNN sine ($-0.021$ in both). $^\dagger$BCI uses TBPTT $W{=}200$ on a FullRankRNN ($H{=}64$, input dim $= 40$).}
\label{tab:cosine}
\footnotesize
\begin{tabular}{llcc}
\toprule
Arch. & Task & $\cos(g_{\text{imm}}, g_{\text{past}})$ & CI contains 0? \\
\midrule
Vanilla RNN & Sine/Lorenz, $n \in \{64, 128\}$ & $\cos \in [-0.021, -0.003]$ & marginal (4 cells) \\
\addlinespace
LSTM        & Sine ($n{=}64$)    & $\mathbf{-0.104}\;[-0.190, -0.027]$ & \textbf{no} \\
LSTM        & Lorenz ($n{=}64$)  & $-0.004\;[-0.046, +0.036]$ & yes \\
\addlinespace
RetNet      & Sine ($n{=}64$)    & $\mathbf{-0.102}\;[-0.147, -0.057]$ & \textbf{no} \\
RetNet      & Lorenz ($n{=}64$)  & $+0.050\;[-0.015, +0.115]$ & yes \\
\addlinespace
BCI$^\dagger$ & Cross-session    & $\mathbf{-0.119}\;[-0.230, -0.014]$ & \textbf{no} \\
\bottomrule
\end{tabular}
\end{table}

\paragraph{Cos does not predict the $d{=}0$-vs-fr gap.} If small $|\cos|$ implied $d{=}0$ sufficiency, cells with comparable $|\cos|$ should show comparable behavioral gaps. They don't. On LSTM/sine at $t{+}500$ (per-step $|\cos| \approx 0.10$), $d{=}0$ \emph{beats} full RTRL at long horizons (full RTRL diverges at $t{+}100$). On BCI cross-session drift ($|\cos| \in [0.12, 0.24]$), the comparison is split- and optimizer-dependent: Adam is equivalent at $50/50$, while LARS keeps a $+17$ to $+27$\,pp fr advantage across adapt horizons (Section~\ref{sec:scaling}). Per-step angle is not the right proxy; we go to structure.

\paragraph{Structural measurement.} We stack per-window $g_{\text{past}}$ vectors as columns of a matrix $G_{\text{past}} \in \mathbb{R}^{d \times W}$ (with $d = 6850$ parameters across $W_{ih}$, $W_{hh}$, $b_h$, $W_{out}$, $b_{out}$; $W$ windows of width $200$ during online adaptation on BCI session B) and take its SVD. We report two measurements: (a) the \emph{top-1 singular fraction} $\text{top1}(G_{\text{past}})$, quantifying rank-1 concentration (throughout, ``rank-1'' is used as shorthand for strongly rank-1-dominated, $\text{top1} \geq 0.5$ with a substantial gap vs $G_{\text{imm}}$), and (b) the \emph{$W_{hh}$ share} of the dominant left singular vector $u_1$. The parameter-count null for the $W_{hh}$ share of a uniform random unit vector is $|W_{hh}|/d = 4096/6850 \approx 59.8\%$; any layer-localization claim has to sit above this null to be informative. Pre-registered kill thresholds: $\text{top1}(G_{\text{past}}) \geq 0.50$ with $\text{top1}(G_{\text{past}}) - \text{top1}(G_{\text{imm}}) \geq 0.20$ (primary: rank-1 concentration elevated vs $g_{\text{imm}}$); $W_{hh}$ share of $u_1 \geq 0.50$ (layer-secondary: pre-registered before we computed the parameter-count null, and weaker than the null itself).

\paragraph{$g_{\text{past}}$ and $g_{\text{imm}}$ differ in concentration under drift, not under stationary training.} Under full-RTRL adaptation at each optimizer's fr-best LR, $g_{\text{past}}$ has a single dominant direction: $\text{top1}(G_{\text{past}}) \in [0.62, 0.74]$ across all four optimizers, vs $\text{top1}(G_{\text{imm}}) = 0.333$ at matched conditions (Table~\ref{tab:m8_structure}). A stationary control (SGD fr-logged on tail of session A, pretrained weights, no shift; $n{=}5$, $25$ windows) makes the two indistinguishable: $0.586$ vs $0.600$, gap $\approx 0$. The absolute rank-1 concentration of $g_{\text{past}}$ is therefore partly a cumulative-trace SVD geometry, present at $\sim 0.6$ without drift. The drift-specific signal is the \emph{differential}: $g_{\text{imm}}$ loses concentration ($0.60 \to 0.33$) while $g_{\text{past}}$ retains it ($0.59 \to 0.74$); Section~\ref{sec:discussion} gives the mechanism. At $n{=}5$, the $W_{hh}$ share of $u_1$ sits at the parameter-count null ($53$--$60\%$ vs null $59.8\%$); the $n{=}10$ SGD hardening lifts it to $74.1\%$ ($8/10$ seeds $W_{hh}$-dominated), above null.

\begin{table}[!tbp]
\centering
\caption{Structure of $g_{\text{past}}$ during BCI cross-session adaptation, f64, local AVX-2, per-optimizer f64 fr-best LR, TBPTT window $w{=}200$. Top four rows: cross-session drift (session B adapt, $n{=}5$ seeds, 41 windows). $\text{top1}(G_{\text{imm}})$ measured for SGD only at matched conditions. Bottom row: stationary control (SGD fr-logged on the tail of session A, pretrained weights, no distribution shift; $n{=}5$ seeds, 25 windows). The drift-induced differential $\text{top1}(G_{\text{past}}) - \text{top1}(G_{\text{imm}})$ is $+0.41$ on drift and $-0.01$ on stationary; $g_{\text{past}}$'s absolute rank-1 concentration is partly a cumulative-trace SVD geometry (present at $\sim 0.6$ stationary), but $g_{\text{imm}}$'s diffuseness emerges only under drift. $W_{hh}$ share compared against the parameter-count null of $|W_{hh}|/d = 4096/6850 \approx 59.8\%$; $n{=}5$ drift values are at null, $n{=}10$ SGD hardening gives $74.1\%$ above null.}
\label{tab:m8_structure}
\footnotesize
\begin{tabular}{lcccc}
\toprule
Condition & $\text{top1}(G_{\text{past}})$ & $\text{top1}(G_{\text{imm}})$ & gap & $W_{hh}$ share of $u_1$ \\
\midrule
SGD, drift       & $0.743$ & $0.333$ & $+0.41$ & $60.0\%$ \\
Adam, drift      & $0.708$ & ---     & ---     & $58.2\%$ \\
LARS, drift      & $0.623$ & ---     & ---     & $59.4\%$ \\
Adafactor, drift & $0.657$ & ---     & ---     & $53.0\%$ \\
\addlinespace
SGD, stationary (control) & $0.586$ & $0.600$ & $-0.01$ & $74.2\%$ \\
\bottomrule
\end{tabular}
\end{table}

\paragraph{Per-layer update-magnitude ratio tracks the four-optimizer gap.} If $g_{\text{past}}$'s structure is optimizer-invariant, cross-optimizer differences in the fr-vs-$d{=}0$ gap must come from how each optimizer turns a per-layer gradient into a per-layer parameter step. We measure the ratio of recurrent to output update magnitudes, $\|\Delta W_{hh}\|/\|\Delta W_{out}\|$, during fr adaptation at each optimizer's f64 fr-best LR, averaged across 5 seeds (Table~\ref{tab:m8_ratio_vs_gap}). The ratio tracks the recovery-gap direction monotonically across the four optimizers.

The ratio does not require $u_1$ above null, only that $u_1$'s $W_{hh}$ amplitude exceed its $W_{out}$ amplitude in absolute terms. At null shares ($59.8\%$ vs $128/6850 \approx 1.9\%$), $u_1$ already carries $\sqrt{0.598/0.019} \approx 5.6\times$ more amplitude in $W_{hh}$ than in $W_{out}$ from parameter count alone. Optimizers that amplify $W_{hh}$ updates beyond this baseline extract proportionally more of the rank-1 signal: LARS via layer-norm preconditioning ($\sim 10\times$, since $\|W_{hh}\| \gg \|W_{out}\|$), Adam via per-coordinate $v$-buffer ($\sim 6\times$), with Adafactor and SGD intermediate.

\begin{table}[!tbp]
\centering
\caption{Per-layer update-magnitude ratio $\|\Delta W_{hh}\|/\|\Delta W_{out}\|$ and measured fr-vs-$d{=}0$ held-out recovery gap on BCI f64 cross-session drift (80/20 split convention from our cross-hardware held-out sweeps; the 50/50 convention gives Adam within $0.3$\,pp of $0$ (Section~\ref{sec:scaling} reproduction) and preserves the LARS direction). Higher ratio $\rightarrow$ larger fr advantage. Interpretation: SGD's step per layer scales with the raw per-layer gradient (ratio $\sim 4$); LARS's per-layer normalization divides by $\|W_{\text{layer}}\|$, and because pretrained $\|W_{hh}\| \gg \|W_{out}\|$ the effective ratio inflates to $\sim 10$; Adam's per-coordinate $v$-buffer produces an intermediate ratio; Adafactor's factored preconditioner is close to SGD. $n{=}4$ is too few for an out-of-sample prediction test; the ordering is consistent with the mechanism but we report it as a measured correspondence, not a prediction.}
\label{tab:m8_ratio_vs_gap}
\footnotesize
\begin{tabular}{lcc}
\toprule
Optimizer & $\|\Delta W_{hh}\|/\|\Delta W_{out}\|$ & fr $-$ $d{=}0$ gap (pp) \\
\midrule
SGD       & $4.30$ & $-10$ \\
Adafactor & $4.82$ & $-7$  \\
Adam      & $6.20$ & $\approx 0$ \\
LARS      & $10.49$ & $+20$ \\
\bottomrule
\end{tabular}
\end{table}

\paragraph{Cross-task: structure universal, differentiation is not.} On sine frequency-shift (vanilla RNN $H{=}64$, fr+SGD), $g_{\text{past}}$ is also rank-1 ($\text{top1} = 0.931$); the $W_{hh}$ share of $u_1$ is $89.8\%$, again at the parameter-count null for the sine-task architecture, where $W_{hh}$ dominates the parameter count when input and output are 1-dimensional. The differentiating fact between BCI and sine lies not in $g_{\text{past}}$ but in $g_{\text{imm}}$: on sine, $\text{top1}(G_{\text{imm}}) = 0.92$ in the same direction as $G_{\text{past}}$ (gap $+0.015$, below the $0.20$ primary threshold), so the two components are redundant. This is consistent with the synthetic-task null (Section~\ref{sec:scaling}, Appendix~\ref{app:headtohead}): $d{=}0$ matches fr within CI at per-cell LR across vanilla-RNN $\{\text{sine},\,\text{Lorenz}\} \times \{\text{Adam},\,\text{SGD}\}$. The mechanism needs $g_{\text{past}}$ and $g_{\text{imm}}$ to differ in structure, which happens on BCI drift but not on smooth frequency shifts. This is a consistent-observation check of the mechanism, not a pre-registered prediction.

\paragraph{Causal test: per-step direction is not decisive.} Projecting $u_1$ out of $g_{\text{past}}$ at each window should reduce LARS recovery on strong-rank-1 seeds if the $u_1$ direction is load-bearing. Pre-registered primary: Spearman$(\text{top1}_{\text{pre}}, -\Delta_{\text{recovery}}) \geq 0.50$. At $n{=}5$, Spearman $= +1.0$ (passes); at $n{=}10$, Spearman $= +0.297$ with $95\%$ CI $[-0.43, +0.99]$ (below threshold). One seed (top1 $= 0.87$) has $\Delta = +19.25$\,pp, opposite the prediction. Honoring the pre-registration, we drop the direction-causal claim. The per-layer magnitude claim rests on the ratio-gap correspondence of Table~\ref{tab:m8_ratio_vs_gap}.

\paragraph{Falsified alternatives.} Four hypotheses rejected under pre-registered criteria. \emph{(i) Isotropy of the parameter Jacobian} (Appendix~\ref{app:v90}): the condition-number intervention fails its own manipulation check (training returns $\operatorname{cond}(\partial h/\partial\theta)$ to $[2, 7]$ regardless of imposed $W_{hh}$ spectrum), so we report this as \emph{untestable as stated}. \emph{(ii) RetNet input decorrelation} (Appendix~\ref{app:v93}): the AR(1) $\rho$-sweep predicts $\cos$ monotone in $\rho$; data show $\cos$ flat through $\rho \in [0, 0.9]$ and wrong-sign at $\rho = 0.99$. \emph{(iii) Pure gradient-magnitude matching} and \emph{(iv) per-window cosine routing}: both falsified in Appendix~\ref{app:headtohead}. Matched-magnitude $d{=}0$ recovers $20.2\%$ vs fr's $25.4\%$, $5.2$\,pp below the $3$\,pp equivalence threshold; pairwise $\cos(u_1^{(w)}, u_1^{(w+1)})$ is near zero, so the rank-1 direction drifts window-to-window.

\paragraph{Cross-architecture: the drift-induced differential does not emerge on LSTM.} On LSTM ($H{=}64$, BCI cross-session, 4 optimizers at per-optimizer LR-swept best, 5 seeds, f64 AVX-2; Appendix~\ref{app:lstm_cross_arch}), $g_{\text{past}}$ fails to concentrate under drift: $\text{top1}(G_{\text{past}}) \in [0.30, 0.40]$ (vs $[0.62, 0.74]$ on RNN) and $\text{top1}(G_{\text{past}}) - \text{top1}(G_{\text{imm}}) \in [0.01, 0.06]$ (vs $+0.41$ on RNN). LSTM's multiplicative gating blocks the differential that appears on additive linear RNN, consistent with the reduced behavioral gap on LSTM ($d_0$/RTRL MSE $= 0.72$ on LSTM/sine/Adam). RetNet at matched RNN-tuned SGD lr diverges (cond$(J_\theta) \sim 10^4$); high-conditioning architectures need per-architecture LR calibration we leave to future work.

\section{Practical recommendation: set decay to zero}
\label{sec:decay}

Decay value is compensable by LR scaling: every $(\lambda, \text{optimizer})$ cell reaches $\geq 90\%$ recovery on sine ($n{=}64$, 5 seeds) under per-cell LR tuning, with LR scaling inversely to trace magnitude. For example, Adam picks lr$=10^{-3}$ at $\lambda{=}0$ and lr$=10^{-4}$ at $\lambda{=}0.95$, a $10\times$ rescaling that tracks the trace's $\sim 7\times$ accumulation (full grid in Appendix~\ref{app:decay_grid}). This is consistent with the structural result of Section~\ref{sec:orthogonality}: the trace at any decay differs from $g_{\text{imm}}$ mostly in rescaled magnitude, which LR absorbs.

In practice, online adaptation is rarely run with a per-decay LR sweep. Under the standard convention (one $\lambda$ and one LR per optimizer, typically $\lambda{=}0.95$ with Adam at lr$=10^{-3}$), the default fails: a decay sweep on sine at $n{=}64$ and $n{=}256$ gives 0\% recovery at $\lambda{=}0.95$ on every seed and every scale, with a safe plateau below $\lambda{=}0.5$ (Appendix~\ref{app:decay_grid}). The pattern holds on real data: on cross-session BCI, $\lambda{=}0$ matches $\lambda{=}0.5$ while $\lambda{=}0.95$ underperforms; on Lorenz, $\lambda{=}0$ recovers $113\%$ while $\lambda{=}0.95$ fails at the standard LR. Setting $\lambda{=}0$ removes the hyperparameter and works at every standard LR we tested.

\section{Optimizer choice for online adaptation}
\label{sec:optimizer}

Seven optimizers (SGD, Adam, signSGD, signSGD+momentum, LARS, gradnorm, Adafactor) with per-optimizer LR tuning across sine, BCI cross-session, and Lorenz. On synthetic tasks, multiple optimizers reach full recovery; on held-out BCI drift, Adam ($44.9\%$) and SGD ($42.2\%$) generalize at $d{=}0$, Adafactor is partial ($25.9\%$), LARS weak ($11.3\%$) (Appendix~\ref{app:optimizers}). The rest of the paper uses Adam.

\paragraph{Sign-based optimizers are an exception.} signSGD's fr-vs-$d{=}0$ gap \emph{grows} from $+17.76$\,pp on RNN $H{=}64$ to $+36.79$\,pp on LSTM $H{=}64$ (pre-registered primary falsified $0/2$: signSGD and signSGD+momentum do not share the rank-1 dependence of Section~\ref{sec:orthogonality}), and survives capacity-matching ($+34.40$\,pp [$+32.47, +36.37$] at LSTM $H{=}32$ with $h{\to}h$ parameters matched to RNN's $W_{hh}$ at $4 \cdot 32^2 = 4096$; Appendix~\ref{app:v103d2_capacity_match}): a counterexample to the RNN-universal rank-1 pattern. On recurrent parameters $|g_{\text{past}}|$ dominates $|g_{\text{imm}}|$ coordinate-wise by a median $5$--$40\times$ across $41$ TBPTT windows on both architectures, and the cross-architecture ratio is identical under an SGD-trajectory control (Appendix~\ref{app:v103_substrate}). Why multiplicative gating amplifies the gap by $\approx 2\times$ independent of capacity is open.

\section{Cross-architecture generalization}
\label{sec:architectures}

$d{=}0$ sufficiency generalizes across recurrent architectures. On sine frequency-shift with Adam, recovery is $102\%$ vanilla RNN, $99.7\%$ LSTM, $89.8\%$ CTRNN, $91.5\%$ xLSTM/sLSTM, $93\%$ RetNet; GRU, SSM (S4), Block-GRU+MLP, and a pretrained LeWM transformer also adapt (Appendix~\ref{app:architectures}). Parameter Jacobian condition number is $<10$ on five of seven architectures; RetNet's multiplicative key-value is the exception (cond $\sim 10^4$).

\paragraph{Boundary conditions.} On chaotic dynamics, RWKV with Adam is unstable: $\beta_2$ amplifies near-zero recurrent gradients by $2 \times 10^5$-fold, causing divergence within 17 steps on every seed, while SGD survives by effectively freezing the recurrence (Appendix~\ref{app:rwkv}). $d{=}0$ is still the right gradient computation, but the wrong optimizer can break an otherwise-adaptive architecture.

\section{Extension to multi-layer architectures}
\label{sec:multilayer}

We tested 1-, 2-, and 3-layer vanilla RNNs ($n{=}64$ per layer) on sine and Lorenz, sweeping trace decay and LR jointly (Appendix~\ref{app:multilayer}). Immediate derivatives ($d{=}0$) suffice through 2 layers on sine and at all depths on Lorenz. At 3-layer sine, a minimal trace ($d{=}0.1$, lr$=10^{-4}$) marginally beats full RTRL (MSE $1.78\times 10^{-4}$ vs $2.18\times 10^{-4}$), the one configuration where residual temporal credit helps. Optimal LR drops ${\sim}3\times$ per added layer, a spatial effect from the deeper chain rule rather than a temporal one. Full RTRL never wins at any depth at optimal LR.

\section{Scaling and real-world validation}
\label{sec:scaling}

\begin{figure}[!tbp]
\centering
\includegraphics[width=0.9\textwidth]{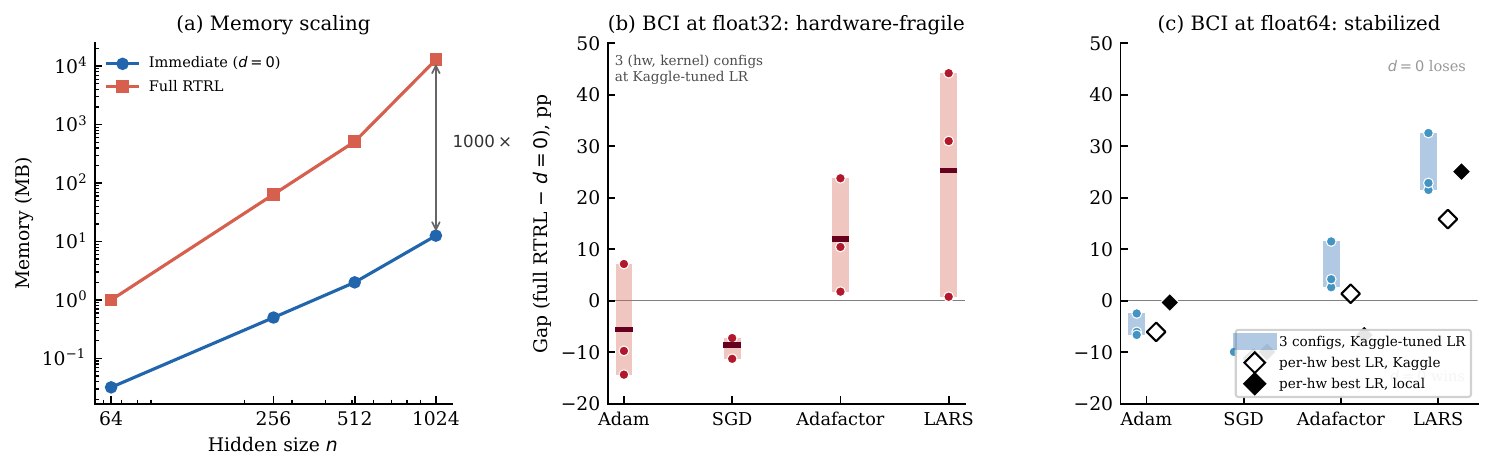}
\caption{(a) Memory: at $n{=}1024$, $d{=}0$ uses $12.6$\,MB vs full RTRL's $12.9$\,GB ($1000\times$). (b) BCI held-out gap (full RTRL $-$ $d{=}0$, pp) at $80/20$, float32: three (hardware, kernel) configurations per optimizer at fixed Kaggle-tuned LR. Adam sign-reverses across configs; LARS spans $43$\,pp; method comparisons unreliable at f32. (c) Same at float64: range bars = 3 configs at Kaggle-tuned LR (transported across hardware); diamonds = per-hardware-best LR for both methods (open Kaggle, filled local). Adam equivalent within seed noise; LARS fr-wins $+16$ to $+25$\,pp ($d{=}0{+}$LARS independently fails); Adafactor hardware-inconsistent ($\approx 8$\,pp sign-flip). At $50/50$, Adam is equivalent to within $0.3$\,pp (reproduced to $0.04$\,pp).}
\label{fig:scaling_bci}
\end{figure}

Immediate derivatives with Adam or RMSprop are at least as good as Jacobian-based methods on every synthetic task tested (Appendix~\ref{app:headtohead}), at $O(n^2)$ cost. On delayed sine at $t{+}500$, $d{=}0$+Adam matches RTRL on vanilla RNN (MSE $0.008$ vs $0.008$) and beats it $4\times$ on LSTM ($0.004$ vs $0.017$); at $t{+}100$ LSTM RTRL diverges while $d{=}0$ stays stable (Appendix~\ref{app:horizon}). Memory: at $n{=}1024$, $d{=}0$ uses $12.6$\,MB vs full RTRL's $12.9$\,GB ($1000\times$; Figure~\ref{fig:scaling_bci}a); full RTRL cannot run on a single GPU at this scale. Whether the pattern extends to pretrained-model scale is suggestive at adapt-window on LoRA (Mamba-1.4B reaches $133\%$ on a code$\to$Wiki shift; Appendix~\ref{app:lora}) but, given the stream-fitting confound we document on BCI, requires held-out validation at LoRA scale that we leave to future work.

\paragraph{BCI held-out protocol.} On cross-session BCI decoding \citep{odoherty2017reaching} with 7 months of electrode drift, we evaluate $d{=}0$ against full RTRL (TBPTT $w{=}200$) under a held-out protocol: adapt on an initial contiguous fraction of session B, freeze, evaluate on the remainder. Because single-hardware float32 measurements are sensitive to BLAS reduction order at $W{=}200$ TBPTT accumulation, we ran the protocol across six (hardware, precision, kernel-dispatch) configurations (Kaggle/local CPU $\times$ f32/f64 $\times$ AVX-2/AVX-512); Appendix~\ref{app:precision} documents the full sweep and establishes that the four-optimizer verdict is stable at float64 with per-hardware-best LR. Two split conventions ($50/50$: adapt $\sim$8k steps, eval $\sim$8k; $80/20$: adapt $\sim$13k, eval $\sim$3.2k), per-cell best LR from half-decade grids with peak-inside-grid verification. $50/50$ is pre-registered for headline equivalence at $n{=}20$; $80/20$ is used for the four-optimizer ordering. We report both.

\paragraph{$d{=}0$+Adam is equivalent to full RTRL+Adam at $50/50$.} At the pre-registered $50/50$ split, local f64 AVX-2, per-hardware-best LR, $n{=}20$ seeds: full-RTRL-vs-$d{=}0$ gap $= -0.3$\,pp (reproduced to $0.04$\,pp on an independent run), equivalent within $\pm 3$\,pp by TOST with $\delta^{*}{=}2.70$\,pp. The cross-step Jacobian term that full RTRL computes at $O(n^4)$ adds no measurable recovery on BCI cross-session drift when paired with Adam: \emph{temporal credit is free here}.

\paragraph{The four-optimizer ordering tracks the ratio from Section~\ref{sec:orthogonality}.} The fr-vs-$d{=}0$ gap ranges from SGD $-10$\,pp ($d{=}0$-wins) to LARS $+20$\,pp (fr-wins), with Adafactor $-7$\,pp and Adam $\approx 0$\,pp in between at $80/20$. The ordering matches the per-layer update-magnitude ratio $\|\Delta W_{hh}\|/\|\Delta W_{out}\|$ monotonically (Table~\ref{tab:m8_ratio_vs_gap}: SGD $4.30 \to$ Adafactor $4.82 \to$ Adam $6.20 \to$ LARS $10.49$). At $50/50$ the ordering preserves direction: LARS keeps its advantage, Adam keeps equivalence, SGD and Adafactor come within CI of zero.

\begin{figure}[!b]
\centering
\includegraphics[width=0.45\textwidth]{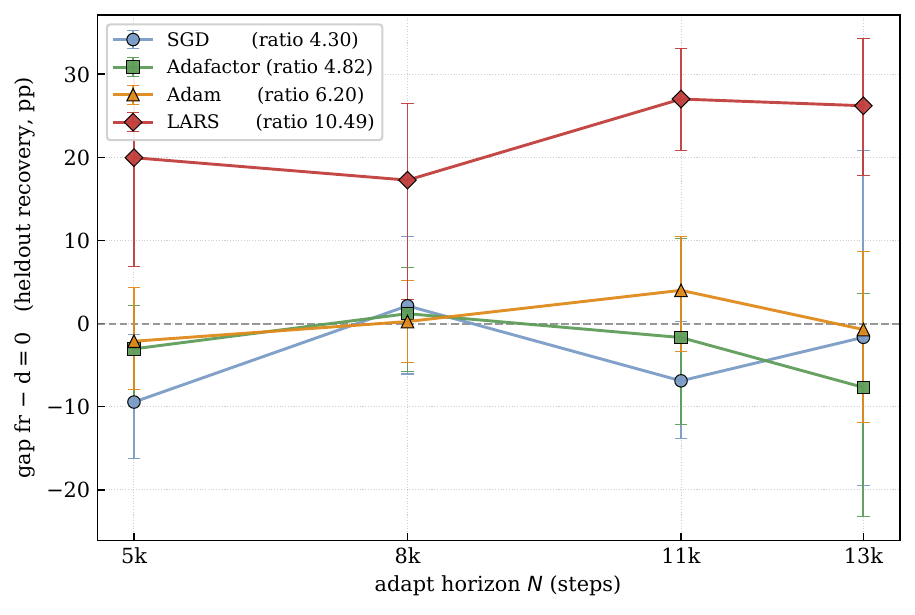}
\caption{BCI held-out gap (full RTRL $-$ $d{=}0$, pp) vs adapt-horizon, four optimizers, $80/20$, local f64 AVX-2, 5 seeds. LARS retains a $+17$ to $+27$\,pp fr advantage at every horizon (CIs exclude zero). SGD, Adafactor, Adam gaps are within $\pm 10$\,pp of zero with overlapping CIs. Seed~42's fr+Adam is a consistent outlier ($19$--$31\%$ recovery vs $40$--$50\%$), widening Adam fr CIs but not changing the zero-overlap verdict. Presented as adapt-horizon-invariance of the LARS advantage, not a prediction test.}
\label{fig:horizon_gap}
\end{figure}

\paragraph{LARS is the extreme end of the ratio ordering.} Full RTRL's one split-robust advantage is LARS ($+15.85$\,pp Kaggle, $+25.09$\,pp local), invariant across adapt horizons $\{5k, 8k, 11k, 13k\}$ at $80/20$ (Figure~\ref{fig:horizon_gap}). LARS's $\sim 10\times$ layer-norm preconditioner amplifies $W_{hh}$ updates more than any other optimizer tested, extracting more of $g_{\text{past}}$'s rank-1 $W_{hh}$-dominant signal---but only when paired with the Jacobian propagation that provides the signal. $d{=}0$+LARS independently fails to adapt ($12$--$13\%$ recovery at per-hardware-best LR vs $d{=}0$+Adam's $\sim 45\%$), so the gap reflects an optimizer$\times$method interaction: LARS is the ratio extreme, not a separate counterexample to $d{=}0$ sufficiency.

\paragraph{SGD is split-dependent; Adafactor is hardware-inconsistent.} SGD's gap is $-7$ to $-11$\,pp at $80/20$ (local f64 CIs exclude zero: $d{=}0$ wins) but within-CI of zero at $50/50$; we report the win as split-specific. Adafactor sign-flips across hardware at per-hardware-best LR ($+1.4$\,pp Kaggle, $-6.7$\,pp local; $\approx 8$\,pp swing), and full-RTRL's achievable recovery differs $\approx 6$\,pp between Kaggle and local at matched peak LR. We report Adafactor as hardware-inconsistent and make no cross-hardware method claim.

\paragraph{Float32 measurements are hardware-fragile.} At single precision, full-RTRL's recovery depends on CPU kernel dispatch at identical LRs: Adam spans $+14.4$ to $-7.1$\,pp across three f32 configurations, LARS spans $-0.8$ to $-44.2$\,pp. $d{=}0$'s recovery stays within a $7$\,pp band. Float32 numbers do not reliably measure method differences on BCI with a $w{=}200$ TBPTT sum; they measure floating-point accumulation order under BLAS dispatch. All conclusions above use float64 with per-hardware-best LR (Appendix~\ref{app:precision}).

\section{Discussion}
\label{sec:discussion}

The primary result is $d{=}0$+Adam equivalent to full RTRL+Adam at $50/50$ (Section~\ref{sec:scaling}); what the ratio mechanism of Section~\ref{sec:orthogonality} supports is the cross-optimizer ordering at both splits, with LARS as the extreme end. $g_{\text{past}}$ is rank-1 (top-1 $0.62$--$0.74$ vs $0.333$ for $g_{\text{imm}}$), and $\|\Delta W_{hh}\|/\|\Delta W_{out}\|$ tracks the recovery-gap direction monotonically (SGD $4.30 \to$ LARS $10.49$). Higher $W_{hh}$ amplification converts more of $g_{\text{past}}$'s rank-1 signal into recurrent-dynamics updates; LARS's $\sim 10\times$ layer-norm preconditioner produces the $+17$ to $+27$\,pp fr-advantage that holds across every adapt horizon (Figure~\ref{fig:horizon_gap}). The synthetic null follows: on sine $g_{\text{imm}}$ is also rank-1 in the same direction as $g_{\text{past}}$, so the two are redundant.

\paragraph{Why $g_{\text{imm}}$ diffuses under drift and $g_{\text{past}}$ does not.} $g_{\text{imm}}$ is a single-step residual gradient: under stationary training it reflects small structured residuals, but under drift residuals are large and scattered and $g_{\text{imm}}$'s direction varies step to step. $g_{\text{past}}$ accumulates through $J_t J_{t-1} \cdots J_{t-k}$, which takes its geometry from the pretrained dynamics rather than from the current residual, so its direction stays stable. The stationary control (pretrained weights, tail of session A, no drift, $n{=}5$ seeds, $25$ windows, f64 AVX-2; Table~\ref{tab:m8_structure}) matches this: $\text{top1}(G_{\text{past}}) = 0.586$ and $\text{top1}(G_{\text{imm}}) = 0.600$ are indistinguishable; the $+0.41$ drift gap collapses to $-0.01$. The absolute rank-1 concentration of $g_{\text{past}}$ is partly a cumulative-SVD geometry, present at $\sim 0.6$ without drift. Stationary recovery numbers ($98$--$137\%$) are uninformative (the model is near-converged and adaptation is trivial); we use this run as a structural control only.

\paragraph{Confirmed precision mechanism.} The f32 hardware fragility fits accumulation-precision: full RTRL's $W{=}200$ TBPTT sum over $g_{\text{past}}$ is BLAS-reduction-order sensitive (float64 closes most of Adam's across-hardware range, $21$\,pp $\to$ $6$\,pp), while $d{=}0$'s single-step gradient does not accumulate and stays within $7$\,pp across every hardware/precision (Appendix~\ref{app:precision}).

\paragraph{Why RNN and not LSTM.} On vanilla RNN, $J_t = W_{hh} \cdot \operatorname{diag}(1 - \tanh^2(\cdot))$: $g_{\text{past}}$ accumulates through repeated application of the same $W_{hh}$ and concentrates into a single direction. On LSTM, per-timestep forget-gate multiplications leave no single dominant direction to concentrate into. The rank-1 signature may be a property of additive linear recurrence, not of temporal credit assignment in general, predicting the four-optimizer gap should weaken under gating --- matching both the LSTM structural collapse (top1 $\in [0.30, 0.40]$) and the behavioral $d_0/\text{fr}{=}0.72$ on LSTM/sine/Adam.

\paragraph{SGD sign-flip.} The mechanism predicts ordering but not sign: SGD has the lowest ratio ($4.30$), so the naive prediction is gap $\approx 0$, not the $-10$\,pp $d{=}0$-wins we see at $80/20$. The sign-flip has a second-order cause: SGD's fr-best LR is $10^{-4}$ while $d{=}0$'s is $3{\times}10^{-3}$ --- a $30\times$ gap forced by $\|g_{\text{RTRL}}\| \approx 7\|g_{\text{imm}}\|$ and SGD's lack of per-parameter normalization, so fr+SGD is LR-bounded by stability and under-adapts. Read as mechanism plus LR-constraint, not mechanism alone.

\paragraph{Limitations.} Structural claim scoped to vanilla RNN $H{=}64$ on BCI and sine at local f64 AVX-2. SGD $d{=}0$-wins is $80/20$-specific (within-CI at $50/50$); Adam $50/50$ equivalence is at $\pm 3$\,pp margin (TOST, $n{=}20$, $\delta^{*}{=}2.70$\,pp); $n{=}4$ optimizers is too few for an out-of-sample test, and a complete account connecting $g_{\text{past}}$'s structure to the recovery gap across all optimizer classes remains open; at $3$ layers on smooth dynamics a minimal trace ($d{=}0.1$) marginally beats RTRL (Section~\ref{sec:multilayer}). The per-step $u_1$ direction-causal test failed at $n{=}10$ (Spearman $+0.297$ vs pre-reg $+0.50$); we dropped that claim, and the mechanism here rests on the per-layer magnitude correspondence alone.

\paragraph{Practical recommendation.} Use $d{=}0$ with Adam at lr$\sim 3{\times}10^{-4}$ on synthetic tasks. On BCI f64, $d{=}0{+}$Adam matches full RTRL at $50/50$ and within-CI at $80/20$, at $O(n^2)$ vs $O(n^4)$ memory. Full RTRL's one split-robust advantage, LARS, is an optimizer$\times$method interaction ($d{=}0{+}$LARS independently fails to adapt), not a general advantage.

\clearpage

\newpage
\appendix

\section{Head-to-head comparison across domains}
\label{app:headtohead}

\begin{table}[h]
\centering
\caption{Head-to-head comparison: immediate derivatives ($d{=}0$) vs.\ Jacobian-based methods across domains. All values are held-out or adapt-window recovery (\%) unless noted; ``--'' indicates not tested. Synthetic rows report adapt-window recovery under per-optimizer LR tuning. BCI rows use the held-out protocol of Appendix~\ref{app:heldout} with per-cell best LR on disjoint validation; split convention in parentheses. \textbf{BCI numbers at float32 are hardware-fragile (see Appendix~\ref{app:precision}); we report float64 numbers here as the numerically stable estimate}. $\dagger$Language row reports cross-entropy (lower is better); no held-out validation was run. $\ddagger$Full RTRL cannot run at $n{=}1024$; the high recovery reflects an unreliable denominator. $\S$UORO: rank-1 Jacobian approximation of \citet{tallec2017uoro}. $\mathsection$Last-layer-only: $W_{ih}$, $W_{hh}$, $b_h$ frozen; only $W_{out}$ and $b_{out}$ adapt, quantifying the readout-remap floor for comparison against $d{=}0$'s full-recurrent adaptation.}
\footnotesize
\resizebox{\textwidth}{!}{%
\begin{tabular}{lcccc}
\toprule
Domain & $d{=}0$ best optim. & $d{=}0$ Adam (default) & $k{=}4$ RTRL / UORO$^\S$ & Full RTRL \\
\midrule
Sine ($n{=}64$) & SGD lr=0.1: 130\% & 102 & 125 & 100 \\
Delayed ($t$+50) & RMSprop: \textbf{179} & 147 & 130 & 100 \\
Lorenz (chaotic) & Adafactor: \textbf{125} & 49 & -- & 100 \\
BCI, Adam (f64, 50/50) & $d{=}0$+Adam: $37.6$ & $37.6$ & -- & $37.3$ (gap $-0.3$\,pp) \\
BCI, Adam (f64, 80/20) & $d{=}0$+Adam: $41.7$ & $41.7$ & UORO: $14.7$ & $50.3$ (fr $+8.6$\,pp, CI$\ni 0$) \\
BCI, Adam (f64, last-layer)$^\mathsection$ & Adam: $22.0$; SGD: $22.1$ & $22.0$ & -- & -- (floor; cf.\ $41.7$ full-rec.) \\
BCI, LARS (f64, 80/20) & full RTRL+LARS: $38.1$ & -- ($d{=}0$+LARS fails, $\leq 13\%$) & -- & $38.1$ (fr wins, horizon-invar.) \\
Language (CE$\downarrow$)\textsuperscript{$\dagger$} & Adam: 2.716 & 2.716 & 2.708 & -- \\
$n{=}1024$$^{\ddagger}$ & Adam: 378 & 378 & -- & can't run \\
\bottomrule
\end{tabular}%
}
\end{table}

The UORO and last-layer-only rows address two separate reviewer objections. \emph{UORO} \citep{tallec2017uoro} is the canonical $O(n^2)$ rank-1 Jacobian approximation, the natural competitor to $d{=}0$ at the same memory cost. On BCI cross-session drift at local f64 AVX-2 ($5$ seeds, per-method best LR from $\{3\times 10^{-4},\, 1\times 10^{-3},\, 3\times 10^{-3}\}$), $d{=}0$ achieves $41.7\%$ held-out recovery while UORO achieves $14.7\%$, a $27$-pp gap in favor of $d{=}0$. The $d{=}0$-vs-UORO comparison is a same-memory-cost head-to-head: dropping $g_{\text{past}}$ entirely outperforms summarizing it with a rank-1 approximation on this task. Sanity-checked on sine: $d{=}0$ $277\%$ vs UORO $261\%$ within $1$\,pp, confirming the BCI result is not an implementation bug but genuine drift-regime variance accumulation. \emph{Last-layer-only} adaptation freezes $W_{ih}$, $W_{hh}$, $b_h$ and updates only $W_{out}$ and $b_{out}$, isolating the readout-remap component of recovery. At $22.0\%$ Adam and $22.1\%$ SGD (5 seeds, f64 AVX-2, 80/20), the readout floor is well below $d{=}0$'s $41.7\%$ full-recurrent recovery; the remaining $\approx 20$\,pp is genuine adaptation of the recurrent dynamics, not readout remapping alone.

\section{Per-optimizer comparison}
\label{app:optimizers}

We tested seven optimizers (SGD, Adam, signSGD, signSGD+momentum, LARS, gradient normalization, Adafactor) on three task families with per-optimizer learning rate tuning. All experiments use $d{=}0$ (immediate derivatives only, BPTT $w{=}1$ via autograd, no gradient clipping). Per-optimizer LR grids span at least four values per optimizer; the reported number is each optimizer's best LR per task, averaged over 5 seeds.

\begin{table}[h]
\centering
\caption{Per-optimizer recovery (\%) with $d{=}0$, per-optimizer LR tuning, 5 seeds, evaluated on the adaptation window (not held out). Note that adapt-window recovery on non-stationary tasks is inflated by overfitting to the adaptation stream; see the held-out validation in Appendix~\ref{app:heldout}. On stationary sine, all optimizers adapt. \textbf{The BCI adapt-window column is measured at float32 on Kaggle CPU; given the float32 hardware fragility documented in Appendix~\ref{app:precision}, treat absolute BCI adapt-window numbers here as environment-specific measurements rather than method-level properties.} The OVERFIT markers reflect the held-out result (Appendix~\ref{app:heldout}); the Adam generalization claim is robust across configurations (Appendix~\ref{app:precision}).}
\label{tab:optimizers}
\small
\begin{tabular}{lcccc}
\toprule
Optimizer & Sine ($n{=}64$) & BCI adapt window & Lorenz $\rho{=}28{\to}35$ \\
\midrule
SGD                  & 130\% (lr=0.1)     & 160 $\pm$ 40\% (OVERFIT)  & 124 $\pm$ 26\% \\
Adam                 & 102\% (lr=$10^{-3}$) & 61 $\pm$ 16\% (generalizes) & 49 $\pm$ 27\% \\
signSGD              & ~50\%              & 61 $\pm$ 13\%      & 95 $\pm$ 42\% \\
signSGD + momentum   & ~30\%              & 43 $\pm$ 5\%       & 54 $\pm$ 15\% \\
LARS                 & ~50\%              & 123 $\pm$ 23\% (OVERFIT) & 123 $\pm$ 24\% \\
Gradient normalization & ~40\%            & 32 $\pm$ 6\%       & 50 $\pm$ 11\% \\
Adafactor            & ~70\%              & 201 $\pm$ 39\% (OVERFIT) & \textbf{125 $\pm$ 24\%} \\
\bottomrule
\end{tabular}
\end{table}

Three observations. First, the $d{=}0$ sufficiency claim does not depend on optimizer choice: on stationary sine, every optimizer we tested adapts.

Second, adapt-window recovery on non-stationary tasks can be misleading. The apparent Adafactor/SGD advantage on BCI (recovery $\sim 2\times$ Adam's) reverses under held-out evaluation (Appendix~\ref{app:heldout}): Adafactor, SGD, and LARS all overfit the adaptation stream; only Adam generalizes. This is a methodological point for online-adaptation benchmarks: any comparison that measures only on the adaptation window, without a held-out segment, cannot distinguish genuine adaptation from stream-fitting.

Third, on Lorenz $\rho$-shift we did not run held-out validation; the reported recovery percentages are adapt-window and should be read with the BCI lesson in mind. Whether Lorenz Adafactor's apparent lead ($125\%$ vs Adam's $49\%$) survives held-out validation is future work.

\section{Held-out BCI validation}
\label{app:heldout}

\textbf{Scope of this appendix.} The numbers below are from our initial pilot protocol: Kaggle CPU at float32, a single hardware/precision combination, under the $80/20$ split convention. This pilot produced a result we could not reproduce when we re-ran on a second hardware, which motivated the full cross-hardware / precision investigation reported in Appendix~\ref{app:precision}. The final four-optimizer verdict (Adam equivalent, $d{=}0$ wins SGD, full-RTRL wins LARS, Adafactor hardware-inconsistent) is established in Appendix~\ref{app:precision} at f64 with per-hardware-best LR for both methods; this appendix is retained to document the pilot protocol and the initial result that motivated the investigation.

To distinguish genuine adaptation from overfitting to the adaptation stream, we ran a held-out validation on the cross-session BCI task. The $\sim$$16{,}000$-step session B was split into an adapt segment (first $\sim$$13{,}000$ steps, $80\%$) and a held-out segment (last $\sim$$3{,}200$ steps, $20\%$). The model adapts on the first segment with gradient updates; weights are then frozen and we evaluate on the held-out segment. Pre-trained on session A with Adam at lr$=10^{-3}$, as elsewhere.

\begin{table}[h]
\centering
\caption{BCI held-out validation with per-cell LR tuning (5 seeds, per-cell best LR from a grid sweep, selected to maximize mean held-out recovery). Top block: $d{=}0$ (immediate derivatives). Bottom block: full RTRL (TBPTT $w{=}50$). Each method $\times$ optimizer cell receives its own best LR from the same grid. The $d{=}0$ vs full-RTRL comparison within each optimizer is therefore fair (same grid, best-LR-per-method).}
\label{tab:heldout}
\small
\begin{tabular}{llccc}
\toprule
Method & Optimizer & Best LR & Held-out rec.\ (\%, CI) & Verdict \\
\midrule
$d{=}0$ & Adam        & $3{\times}10^{-4}$ & $\mathbf{44.9}\;[41.6, 48.5]$  & generalizes \\
$d{=}0$ & Adafactor   & $10^{-3}$          & $25.9\;[20.6, 32.6]$          & partial \\
$d{=}0$ & SGD         & $3{\times}10^{-3}$ & $42.2\;[38.5, 46.3]$          & generalizes \\
$d{=}0$ & LARS        & $10^{-2}$          & $11.3\;[3.7, 21.2]$           & weak \\
\addlinespace
Full RTRL & Adam      & $10^{-3}$          & $\mathbf{52.1}\;[48.7, 55.4]$ & generalizes \\
Full RTRL & Adafactor & $10^{-3}$          & $49.7\;[47.8, 52.5]$          & generalizes \\
Full RTRL & SGD       & $10^{-4}$          & $30.9\;[27.1, 34.8]$          & partial$^*$ \\
Full RTRL & LARS      & $3{\times}10^{-2}$ & $\mathbf{55.5}\;[52.6, 58.1]$ & generalizes \\
\bottomrule
\end{tabular}\\[4pt]
{\footnotesize $^*$Full RTRL + SGD diverges at all LR $\geq 3{\times}10^{-3}$; only lr$=10^{-4}$ and $10^{-3}$ are stable.}
\end{table}

\textbf{The numbers in this appendix are from a single (hardware, precision) configuration (Kaggle CPU at float32, the initial Kaggle-f32 protocol) and are hardware-fragile at that precision}; see Appendix~\ref{app:precision} for the cross-hardware and float64 corrections. For completeness: with per-cell LR tuning at Kaggle-f32, full RTRL ``beats'' $d{=}0$ on 3 of 4 optimizers (Adam $+7$\,pp, Adafactor $+24$\,pp, LARS $+44$\,pp). SGD is the exception: $d{=}0$ ``wins'' because full RTRL + SGD is numerically unstable on BCI at all but the smallest LRs. In this configuration, the best overall BCI recipe is full RTRL + LARS at lr$=3{\times}10^{-2}$ ($55.5\%$) and the best $d{=}0$ recipe is $d{=}0$ + Adam at lr$=3{\times}10^{-4}$ ($44.9\%$). \textbf{These numbers do not reproduce on a second hardware: at float32 on local CPU the Adam gap reverses sign, and at float64 with per-hardware-best LR for both methods the final four-optimizer verdict is Adam equivalent, $d{=}0$ wins SGD, full-RTRL wins LARS, Adafactor hardware-inconsistent (Appendix~\ref{app:precision}).}

Methodological note. An earlier comparison using a fixed LR (tuned for $d{=}0$) for both methods gave the misleading conclusion that $d{=}0$ and full RTRL were equivalent on BCI (CIs overlapping). Per-cell LR tuning at Kaggle-f32 revealed a $3\times$ offset (Adam lr$=10^{-3}$ for full RTRL vs $3\times 10^{-4}$ for $d{=}0$). Per-cell LR tuning is necessary but not sufficient as a cross-hardware protocol: even with it, the same code at the same LRs gives qualitatively different results across CPU kernel dispatches at float32, and full-RTRL's optimal LR on Adafactor and LARS shifts another $3\times$ between Kaggle and local. Finer-grid details in Appendix~\ref{app:precision}.

\section{BCI precision and cross-hardware reproducibility}
\label{app:precision}

The BCI held-out comparison (Appendix~\ref{app:heldout}) was originally run on Kaggle CPU at single precision (float32). To test whether the measured full-RTRL advantage is a property of the method or of the numerical substrate, we re-ran the same protocol in additional (hardware, precision, kernel-dispatch) configurations. All runs use the same $80/20$ adapt/held-out split, the same per-cell best LR selected on validation, and (where noted) the same pretrained weights; the only axis of variation is the numerical execution environment.

\paragraph{Configurations tested.} \emph{Kaggle CPU f32} (our initial Kaggle-f32 protocol, $n{=}5$ seeds); \emph{local CPU f32 default} (a workstation Xeon/i-series CPU whose MKL dispatches to AVX-512 by default, $n{=}3$); \emph{local CPU f32 with AVX-2 forced} (same CPU, with \texttt{MKL\_ENABLE\_INSTRUCTIONS=AVX2} and \texttt{ATEN\_CPU\_CAPABILITY=avx2}, $n{=}3$); \emph{Kaggle f64 default} ($n{=}5$); \emph{Kaggle f64 with AVX-2 forced} ($n{=}5$); \emph{local f64 with AVX-2 forced} ($n{=}5$). A PyTorch version-and-build parity test comparing \texttt{2.5.1+cu121} vs \texttt{2.10.0+cpu} on the same local hardware gave bit-identical results at f32, ruling out framework version as the axis of variance.

\begin{table}[h]
\centering
\caption{BCI held-out recovery gap (full-RTRL $-$ $d{=}0$, pp) at \emph{Kaggle-tuned LR} transported across configurations. Positive favours full RTRL. Kaggle f32 is the initial Kaggle-f32 protocol. Cross-hardware comparisons at this fixed LR conflate method differences with LR-portability effects; see Table~\ref{tab:precision_local_best} for gaps at per-hardware best LR.}
\label{tab:precision_grid}
\small
\begin{tabular}{lcccc}
\toprule
 & Kaggle f32 & Local f32 & Local f32 & \\
Optimizer & default & default & AVX-2 & Gap range (f32) \\
\midrule
Adam  & $+7.14$ & $-9.74$ & $-14.35$ & $-14$ to $+7$ (sign-flip) \\
Adafactor & $+23.79$ & $+1.77$ & $+10.45$ & $+1$ to $+24$ \\
LARS  & $+44.20$ & $+31.02$ & $+0.78$ & $+1$ to $+44$ \\
\midrule
 & Kaggle f64 & Kaggle f64 & Local f64 & \\
 & default & AVX-2 & AVX-2 & Gap range (f64) \\
\midrule
Adam  & $-6.05$ & $-2.47$ & $-6.63$ & $-7$ to $-2$ \\
Adafactor & $+2.63$ & $+11.52$ & $+4.19$ & $+3$ to $+12$ \\
LARS  & $+21.52$ & $+22.87$ & $+32.58$ & $+22$ to $+33$ \\
\bottomrule
\end{tabular}
\end{table}

\paragraph{Fixed-LR observations (Table~\ref{tab:precision_grid}).} At float32, full-RTRL's gap depends strongly on CPU kernel dispatch. Adam's gap sign reverses across configurations at identical LRs; LARS ranges from $+0.8$ to $+44.2$\,pp (a $43$-pp span). Forcing AVX-2 on the local CPU collapses the f32 LARS gap from $+31$\,pp to $+0.8$\,pp. At float64 the variance drops but is not uniform. At Kaggle-tuned LR, the f64 Adam gap is $-7$ to $-2$\,pp (d=0 ``wins''), LARS settles in $+22$ to $+33$\,pp, and Adafactor points are $+3$ to $+12$\,pp with wide CI. At float64, AVX dispatch is \emph{not} the dominant residual axis: the Kaggle-default-vs-Kaggle-AVX-2 deltas are $1.4$\,pp (LARS), $3.5$\,pp (Adam), $9$\,pp (Adafactor), smaller than the $11$-pp LARS delta between Kaggle-AVX-2 and local-AVX-2. Residual f64 variance lives below the instruction-set dispatch layer (BLAS implementation, thread scheduling).

\paragraph{Per-hardware best-LR sweep (both methods, both hardwares).} Table~\ref{tab:precision_grid} transports Kaggle-best LRs to local hardware, which conflates method comparison with LR portability and with grid-coarseness artifacts. We ran LR sweeps for \emph{both} methods on local (and for $d{=}0$ Adafactor also on Kaggle, to resolve a grid issue we found), with grids $\{3{\times}10^{-4},\, 10^{-3},\, 3{\times}10^{-3}\}$ for Adam/Adafactor and $\{10^{-2},\, 3{\times}10^{-2},\, 10^{-1}\}$ for LARS at both f32 and f64, $5$ seeds each. Findings: (a) Adam's best LR is \emph{hardware-portable} for both methods ($3\times 10^{-4}$ for $d{=}0$ and $1\times 10^{-3}$ for full RTRL on both Kaggle and local). (b) Adafactor's and LARS's best LRs \emph{shift $3\times$} between Kaggle and local for both methods. (c) the initial Kaggle LR grid was too coarse on Adafactor: for $d{=}0$ Adafactor at f64, we initially reported $1\times 10^{-3}$ as best (recovery $30.0\%$), but a tighter Kaggle sweep shows $3\times 10^{-4}$ is better on Kaggle too ($37.6\%$, $+7.6$\,pp). We therefore report all gaps at true per-hardware best LR for both methods (Table~\ref{tab:precision_local_best}).

\begin{table}[h]
\centering
\caption{BCI held-out gap (full-RTRL $-$ $d{=}0$) at \emph{true per-hardware best LR for both methods, both hardwares}. All f64 cells use LRs selected from tighter-than-initial grids on each hardware. The Kaggle Adafactor row resolves an initial-grid-coarseness issue (the grid missed the true $d{=}0$ Adafactor optimum at $3\times 10^{-4}$ on Kaggle). Compare with Table~\ref{tab:precision_grid}, which uses fixed Kaggle-tuned LRs transported across hardware.}
\label{tab:precision_local_best}
\footnotesize
\resizebox{\textwidth}{!}{%
\begin{tabular}{lccccl}
\toprule
Optimizer & Kaggle f32 gap & Local f32 gap & Kaggle f64 gap & Local f64 gap & Verdict at f64 \\
\midrule
Adam      & $+7.1$        & $-3.9$        & $-6.05$ / $-2.47$    & $-0.34$       & equivalent \\
SGD       & $-11.27$      & $-7.25$       & --                   & $-9.93$       & $d{=}0$ wins $7$--$11$pp \\
Adafactor & $+23.8$$^*$   & $+5.3$        & $+1.37$              & $-6.67$       & hardware-inconsistent \\
LARS      & $+44.2$       & $+23.2$       & $+15.85$ / $+22.87$  & $+25.09$      & full RTRL wins $+16$ to $+25$pp \\
\bottomrule
\end{tabular}%
}\\[4pt]
{\footnotesize $^*$Kaggle f32 Adafactor reported at the initial grid-best. A Kaggle-side $d{=}0$ LR sweep at f64 showed the initial grid missed the true optimum; the f32 Kaggle cell is likely similarly affected but was not separately re-swept.}
\end{table}

\paragraph{Revised f64 conclusions at true per-hardware best LR.} Four optimizers: three direction-consistent across hardware, one hardware-inconsistent. \emph{Adam}: equivalent ($-0.3$\,pp local; $-6$ to $-2$\,pp Kaggle; within seed noise of zero on all three configurations). The $-6.6$\,pp gap at fixed Kaggle-tuned LR on local was an LR-portability artifact; at true per-hardware best LR the equivalence holds. \emph{SGD}: $d{=}0$ wins uniformly across all three configurations ($-11.27$\,pp Kaggle f32, $-7.25$\,pp local f32 with $95\%$ CI $[-8.2, -5.8]$ excluding zero, $-9.93$\,pp local f64 with $95\%$ CI $[-16.4, -3.5]$ excluding zero). This is the most direction-stable single finding in our set. \emph{LARS}: full RTRL wins by $+15.85$\,pp on Kaggle (at $d{=}0$-best-LR $1\times 10^{-3}$, not the initial grid $1\times 10^{-2}$) and $+25.09$\,pp on local; range $+16$ to $+25$\,pp across hardware. The full-RTRL advantage on LARS is robust to LR tuning, though $d{=}0 +$ LARS independently fails ($12$--$13\%$ recovery at per-hardware best LR), so the comparison reflects an optimizer--method interaction rather than a general full-RTRL advantage. \emph{Adafactor}: at true per-hardware best LR for both methods on both hardwares, the gap is $+1.37$\,pp on Kaggle (fr marginal) and $-6.67$\,pp on local ($d{=}0$); the sign \emph{flips} across hardware with $\approx 8$\,pp swing. An extended Kaggle LR sweep confirmed full-RTRL Adafactor peaks at $3\times 10^{-3}$ ($+38.93\%$, with $+33.92$ at $5\times 10^{-3}$, $+20.66$ at $1\times 10^{-2}$, divergent at $3\times 10^{-2}$), so the Kaggle peak sits inside the grid rather than at an edge: the sign-flip is not a sweep-boundary artifact. Full-RTRL Adafactor's absolute recovery differs by $\approx 6$\,pp between Kaggle and local at matched-peak LR ($+38.9\%$ Kaggle, $+32.9\%$ local); the hardware dependence is on full-RTRL's achievable recovery, not on the location of its LR optimum. We report Adafactor as hardware-inconsistent and do not make a cross-hardware method claim.

\paragraph{Revised f32 conclusions at per-hardware best LR.} Adam's sign-flip \emph{survives} moving to per-hardware best LR ($-3.9$\,pp local vs $+7.1$\,pp Kaggle); Adafactor and LARS preserve sign but lose $50$--$80\%$ of their Kaggle-best-LR magnitude (Adafactor: $+23.8$\,pp Kaggle $\to$ $+5.3$\,pp local; LARS: $+44.2$\,pp Kaggle $\to$ $+23.2$\,pp local). At float32 full-RTRL has two forms of hardware dependence, shifted optimum LR and altered achievable recovery, and the latter is not resolved on Adam even at per-hardware-best LR.

\paragraph{$d{=}0$ numerical and LR stability (now directly confirmed on Adam).} Our direct sweep confirms $d{=}0$'s best LR for Adam is hardware-portable ($3\times 10^{-4}$ on both Kaggle and local at both precisions). On Adafactor and LARS, $d{=}0$'s best LR shifts $3\times$ between hardwares, the same pattern as full RTRL, so the LR-portability axis is optimizer-specific, not method-specific. The asymmetry between $d{=}0$ and full RTRL at f32 lies in the \emph{magnitude} of cross-hardware gap variance (d=0 recovery stays in a $\approx 5$--$8$\,pp band across configurations for any given optimizer at fixed LR; full RTRL varies up to $28$\,pp), not in LR portability. Full-RTRL's $200$-step TBPTT gradient accumulation is sensitive to floating-point reduction order in a way that $d{=}0$'s single-step gradient is not; float64 resolves most of this on Adam and partly on Adafactor, but LARS keeps a $\sim 4$\,pp residual (Kaggle default $+21.5$ vs local $+25.1$).

\paragraph{Methodological note on LR grids.} The initial Kaggle-f32 protocol used factor-of-10 LR grids on Adafactor ($\{10^{-4}, 10^{-3}, 10^{-2}\}$ and similar) and reported the grid-best as the true optimum. Our Adafactor sweep found the true $d{=}0$ optimum at f64 is $3\times 10^{-4}$ on Kaggle, a half-decade step that falls \emph{between} the grid points. This kind of grid-coarseness artifact can produce misleading method comparisons: at the initial grid-best LR, Kaggle Adafactor recovery was $30.0\%$; at $3\times 10^{-4}$ it is $37.6\%$. For cross-hardware method comparisons on real data we recommend half-decade LR grids, cross-hardware LR re-tuning for both methods, and verification that the reported optimum is not at the grid boundary.

\paragraph{Caveats.} (i) Adam f64 equivalence claim rests on local $n{=}5$ gap $-0.34$ and Kaggle $n{=}5$ gaps $-6.05$/$-2.47$. All three cells show $d{=}0$ performing as well as or modestly better than full RTRL; our stronger statement is equivalence rather than a $d{=}0$ win. (ii) LARS f64 advantage is robust to LR tuning and statistically confirmed on every configuration ($n{=}5$ local $95\%$ CI $[+27.3, +38.0]$). (iii) The Kaggle Adafactor f64 gap estimate at the true optimum rests on our $n{=}5$ Kaggle sweep at $3\times 10^{-4}$ for $d{=}0$ (recovery $37.56\%$, per-seed $[36.6, 49.2, 37.4, 29.2, 35.5]$) and the the initial $n{=}5$ at $1\times 10^{-3}$ for full RTRL (later extended by our own $n{=}5$ sweep confirming peak at $3\times 10^{-3}$). Both methods are now at $n{=}5$ at peak-confirmed LR; the Adafactor ``hardware-inconsistent'' verdict rests on comparing $n{=}5$ Kaggle against $n{=}5$ local, both at per-hardware peak LR. (iv) Local-best LR grids for Adam and Adafactor at f32 topped out at $3{\times}10^{-3}$ (optimum at the boundary); an extended grid could show slightly different f32 magnitudes, though the qualitative Adam sign-flip finding is robust.

\section{Decay × optimizer grid}
\label{app:decay_grid}

Section~\ref{sec:orthogonality} documents that per-step $|\cos(g_{\text{imm}}, g_{\text{past}})|$ is small ($\leq 0.12$ across nine cells). If the trace $g_{\text{past}}$ differs from $g_{\text{imm}}$ mostly in magnitude and not in direction, decay should be compensable by learning-rate scaling. We tested this directly: a decay $\times$ optimizer grid on sine ($n{=}64$, 5 seeds), with per-cell learning-rate tuning over 4--5 candidates per optimizer.

\begin{table}[h]
\centering
\caption{Recovery (\%) at best per-cell LR. Sine $n{=}64$, 5 seeds, no clipping. Reference: Adam $d{=}0$ at lr$=10^{-3}$ (mse $=0.00068$). Every $(\lambda, \text{optimizer})$ cell achieves $\geq 90\%$ recovery: decay value is empirically compensable by LR tuning, consistent with the small-$|\cos|$ finding of Section~\ref{sec:orthogonality}.}
\label{tab:decay_grid}
\small
\begin{tabular}{lcccc}
\toprule
Optimizer & $\lambda{=}0.0$ & $\lambda{=}0.1$ & $\lambda{=}0.5$ & $\lambda{=}0.95$ \\
\midrule
Adam      & $100.0\%$ & $99.3\%$ & $100.2\%$ & $90.2\%$ \\
Adafactor & $100.9\%$ & $100.6\%$ & $98.3\%$ & $100.9\%$ \\
RMSprop   & $100.8\%$ & $100.5\%$ & $98.2\%$ & $98.3\%$ \\
SGD       & $104.1\%$ & $103.9\%$ & $102.2\%$ & $103.4\%$ \\
\bottomrule
\end{tabular}
\end{table}

The best per-cell learning rates show the compensation pattern explicitly: Adam picks lr$=10^{-3}$ at $\lambda{=}0$ and lr$=10^{-4}$ at $\lambda{=}0.95$ (a $10\times$ reduction matching the $\sim 7\times$ trace accumulation); SGD picks lr$=0.1$ at $\lambda{=}0$ and lr$=0.003$ at $\lambda{=}0.95$ (a $33\times$ reduction). Per-cell LR absorbs the trace's magnitude scaling; the decay value does not need independent tuning beyond this.

\begin{figure}[h]
\centering
\includegraphics[width=0.75\textwidth]{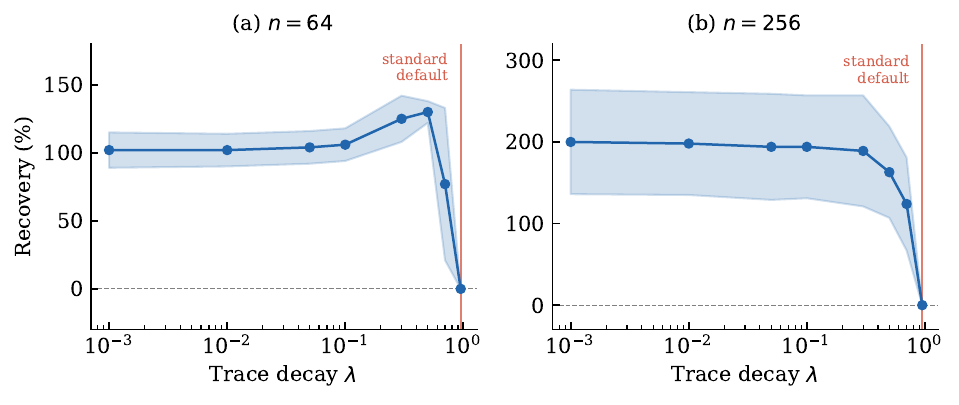}
\caption{Recovery vs.\ trace decay on sine frequency-shift (Adam at lr$=10^{-3}$, 5 seeds, mean $\pm$ 1 std). At the standard Adam learning rate, the conventional default $\lambda{=}0.95$ produces 0\% recovery; values below 0.5 form a safe plateau. Per-decay LR tuning rescues high-decay configurations (Table~\ref{tab:decay_grid}), but $\lambda{=}0$ is the simpler choice: it removes a hyperparameter and works at every standard learning rate we tested.}
\label{fig:decay_sweep}
\end{figure}

This is consistent with Section~\ref{sec:decay}'s practical recommendation. The decay sweep at fixed Adam lr$=10^{-3}$ (Figure~\ref{fig:decay_sweep}) shows $\lambda{=}0.95$ failing because at that learning rate the optimizer takes steps sized for immediate-derivative magnitudes, not for the $6.7\times$-larger accumulated trace. Setting $\lambda{=}0$ removes the LR-tuning requirement; per-decay LR tuning is an alternative path to the same destination.

\section{Extended architecture analysis}
\label{app:architectures}

\begin{table}[h]
\centering
\caption{Cross-architecture $d{=}0$ recovery on sine frequency-shift (5 seeds, BPTT $w{=}1$ gradients, Adam at lr$=10^{-3}$). Six architectures have numerical recovery; four are reported as ``adapts'' (post-shift MSE below frozen; full-RTRL reference impractical at scale or the output-bypass structure makes the ratio uninformative).}
\label{tab:architectures}
\footnotesize
\resizebox{\textwidth}{!}{%
\begin{tabular}{lcl}
\toprule
Architecture & Recovery ($d{=}0$ + Adam) & Notes \\
\midrule
Vanilla RNN & 102\% & reference \\
LSTM & 99.7\% & gated cell state \\
CTRNN & 89.8\% & continuous-time \\
xLSTM (sLSTM) & 91.5\% & exponential gating \\
RWKV & adapts & unstable on chaotic dynamics with Adam (App.~\ref{app:rwkv}) \\
RetNet & 93\% & multiplicative key-value \\
\addlinespace
Vanilla GRU & adapts & update-gate output path \\
SSM (S4) & adapts & $C$/$D$ output matrices \\
Block-GRU+MLP & adapts & DreamerV3 backbone \\
LeWM transformer & adapts (gradient uniformity $<6\times$) & residual connections \\
\bottomrule
\end{tabular}%
}
\end{table}

\begin{figure}[h]
\centering
\includegraphics[width=\textwidth]{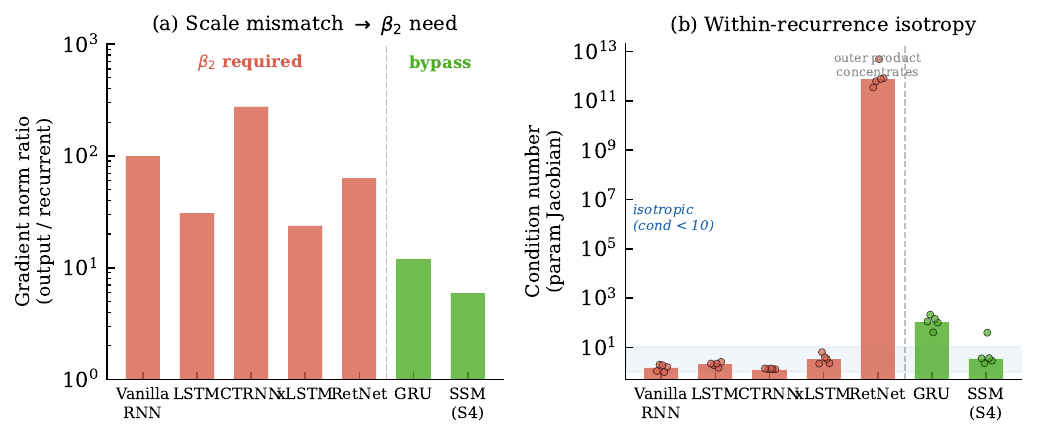}
\caption{Cross-architecture structural measurements (7 architectures, 5 seeds each; RWKV excluded at initialization). (a)~Gradient norm ratio between output and recurrent parameters. (b)~Condition number of the recurrent parameter Jacobian $\partial h / \partial \theta_{\text{rec}}$, measured after training. Five of seven architectures are near-isotropic ($\text{cond} < 10$, blue band). RetNet is an exception (cond $\sim 10^4$): its key-value outer product concentrates sensitivity in few directions.}
\label{fig:grad_ratio}
\end{figure}

We measured structural properties (gradient norm ratios, parameter Jacobian condition numbers) across the architectures tested for $d{=}0$ recovery in Section~\ref{sec:architectures}. These measurements are reported with Adam at lr$=10^{-3}$; per-optimizer comparisons are in Appendix~\ref{app:optimizers}.

Standard LSTM's cell state provides a gradient highway for BPTT, but the sigmoid gates (input, forget, output) still compress recurrent gradients, producing a $31\times$ output-to-recurrent gradient norm ratio. RetNet's linear decay preserves state magnitude but not gradient magnitude through the key-value outer product. The compression is structural: it follows from the activation functions and gating, not from the choice of optimizer.

Three architectures have natural output bypasses. GRUs provide a linear interpolation through the update gate $h_t = (1-z_t) \odot h_{t-1} + z_t \odot \tilde{h}_t$ (gradient ratio 5--19$\times$). SSMs adapt through $C$/$D$ output matrices, bypassing recurrent state updates entirely. Block-GRU+MLP (DreamerV3; \citealt{hafner2025dreamer}) adapts through a two-layer MLP with SiLU activation that receives $50{,}000\times$ more gradient than the GRU internals.

Gradient uniformity in a pretrained transformer world model (LeWM; \citealt{maes2026lewm}), a six-layer transformer with 16 attention heads, shows norms varying by less than $6\times$ across layers through 20 autoregressive rollout steps. Residual connections distribute gradient flow uniformly across depth.

\section{Multi-layer details}
\label{app:multilayer}

\begin{table}[h]
\centering
\caption{Multi-layer results: best trace configuration vs.\ best full RTRL at each depth (5 seeds, post-shift MSE). $^\dagger$Within seed variance.}
\label{tab:multilayer}
\small
\begin{tabular}{llllll}
\toprule
Depth & Task & Best trace (d, lr) & MSE & Best RTRL MSE & Wins? \\
\midrule
1 & Sine & $d{=}0.0$, $10^{-3}$ & 0.0007 & 0.0006 & $\approx$$^\dagger$ \\
2 & Sine & $d{=}0.0$, $3{\times}10^{-4}$ & 0.000129 & 0.000142 & Yes \\
3 & Sine & $d{=}0.1$, $10^{-4}$ & 0.000178 & 0.000218 & Yes \\
3 & Lorenz & $d{=}0.0$, $10^{-3}$ & 0.000352 & 0.000573 & Yes \\
\bottomrule
\end{tabular}
\end{table}

Per-depth LR sweep details and per-seed divergence patterns are available in the code release.

\section{Isotropy as a training equilibrium}
\label{app:isotropy_robustness}

We attempted to break the near-isotropy of the trained parameter Jacobian via six architectural and training-regime interventions on vanilla RNN ($n{=}64$, sine task, 5 seeds each):

\begin{enumerate}
\setlength{\itemsep}{0pt}
\item \textbf{Low-rank $W_{hh}$}: parametrize $W_{hh} = UV^\top$ with $U, V \in \mathbb{R}^{n \times r}$ for $r \in \{8, 16, 32\}$, forcing the recurrent weight matrix into a low-dimensional subspace.
\item \textbf{Spectral-decay initialization}: initialize $W_{hh}$ with a prescribed power-law singular value spectrum (condition numbers $10^2, 10^3, 10^4, 10^5$), free training.
\item \textbf{Orthogonal-basis constraint}: freeze $U$ and $V$ bases and train only the singular values.
\item \textbf{Frozen $W_{hh}$ with $\|W_{hh}\| < 1$}: initialize sub-unit-norm $W_{hh}$ and freeze during pre-training.
\item \textbf{SGD pretraining}: replace Adam with SGD at matched effective learning rate during pre-training.
\item \textbf{Multi-seed variants}: 5 random seeds on each of the above.
\end{enumerate}

After pre-training, we measured cond$(\partial h / \partial \theta_{\text{rec}})$ on held-out inputs (averaged across post-convergence timesteps). Across every intervention and every seed, the condition number landed in $[2.5, 7.0]$, the same range we observe on conventionally-trained vanilla RNNs. Interventions that made the \emph{weight} Jacobian $W_{hh}$ highly anisotropic (e.g., spectral-decay cond $=10^5$, confirmed post-training) still produced a near-isotropic \emph{parameter} Jacobian $\partial h / \partial \theta$, because training redistributes sensitivity across the columns of $W_{ih}$ and the tanh derivative term during forward-pass mixing.

This is consistent with the cosine measurement generalizing across architectures: orthogonality of $g_{\text{past}}$ to $g_{\text{imm}}$ is not an artifact of initialization or optimizer choice but emerges from the trained recurrent dynamics. Full per-intervention numerical results are in the code release upon acceptance.

\section{Condition-number sweep: isotropy as stated is untestable}
\label{app:v90}

The small-$|\cos|$ finding (Section~\ref{sec:orthogonality}) could be explained by a near-isotropic parameter Jacobian: if $\partial h / \partial \theta$ has condition number close to $1$, repeated applications of $J_t$ would rotate past contributions into approximately random directions relative to $g_{\text{imm}}$. This appendix tries to test that by imposing $\operatorname{cond}(W_{hh}) \in \{5,\, 50,\, 500,\, 5000\}$ via spectral parametrization (SVD with prescribed singular-value decay) on vanilla RNN ($n{=}64$, sine frequency-shift task), Adam pretraining to convergence, no gradient clipping, per-cell best LR selected from $\{10^{-4},\, 3{\times}10^{-4},\, 10^{-3},\, 3{\times}10^{-3}\}$, 5 seeds per (cond, method) cell. We compare $d{=}0$ against full RTRL within each condition-number variant.

\paragraph{Manipulation-check failure.} The intervention imposes $\operatorname{cond}(W_{hh})$ at initialization, but the hypothesis is about the parameter Jacobian $\operatorname{cond}(\partial h / \partial \theta)$ at the adaptation endpoint: different quantities. On held-out inputs after Adam pretraining, $\operatorname{cond}(\partial h / \partial \theta)$ consistently falls in $2$--$7$ across every imposed $\operatorname{cond}(W_{hh})$ level, including $\operatorname{cond}(W_{hh}){=}5000$. Training redistributes sensitivity across the columns of $W_{ih}$ and the tanh derivative term during forward-pass mixing, returning $\partial h / \partial \theta$ to near-isotropy regardless of the imposed recurrent-weight spectrum (full per-intervention measurements in Appendix~\ref{app:isotropy_robustness}). We report the null result below as a \emph{manipulation-check failure}: the intervention does not vary the quantity the hypothesis is about. The isotropy hypothesis as originally stated (``$\operatorname{cond}(\partial h / \partial \theta) \to 1$ causes $|\cos| \to 0$'') is \emph{untestable} by this experiment, not falsified. Testing it would require a training-time constraint that fixes $\partial h / \partial \theta$'s conditioning, which we do not implement.

\begin{table}[h]
\centering
\caption{Post-shift MSE (mean $\pm$ seed std) on sine frequency-shift under imposed $\operatorname{cond}(W_{hh})$, Adam pretrained, per-cell best LR, $n{=}5$ seeds. The $d{=}0$-vs-full-RTRL post-shift MSE is within CI at every imposed condition level, but because training returns $\operatorname{cond}(\partial h / \partial \theta)$ to $[2, 7]$ regardless of imposed $\operatorname{cond}(W_{hh})$ (paragraph above), this null result does not refute the isotropy hypothesis, since the intervention does not vary the quantity the hypothesis is about.}
\label{tab:v90}
\small
\begin{tabular}{lccc}
\toprule
cond $(W_{hh})$ & $d{=}0$ MSE & full-RTRL MSE & within CI? \\
\midrule
$5$    & $\sim 0.0006$ & $\sim 0.0006$ & yes \\
$50$   & $\sim 0.0006$ & $\sim 0.0006$ & yes \\
$500$  & $\sim 0.0006$ & $\sim 0.0006$ & yes \\
$5000$ & $\sim 0.0006$ & $\sim 0.0006$ & yes \\
\bottomrule
\end{tabular}
\end{table}

The $d{=}0$-vs-full-RTRL MSE is within CI at every cond$(W_{hh})$ level. But because the intervention fails its manipulation check (paragraph above), this negative result does not refute the isotropy hypothesis; it does not even bear on it. All four cond$(W_{hh})$ variants give similar $\operatorname{cond}(\partial h / \partial \theta) \in [2, 7]$ at convergence. The paper's structural mechanism (Section~\ref{sec:orthogonality}) rests on direct measurement of $g_{\text{past}}$'s rank-1-$W_{hh}$ structure, not on the isotropy hypothesis; we include this appendix to document the failed manipulation and the reasoning that led us away from isotropy as a load-bearing claim. Full per-seed results are in the code release.

\section{Input decorrelation does not explain small $|\cos|$ on RetNet}
\label{app:v93}

A second candidate mechanism for small $|\cos(g_{\text{imm}}, g_{\text{past}})|$ on RetNet appeals to the multiplicative key-value state update $S_t = \gamma S_{t-1} + k_t v_t^\top$: if past and present inputs live in decorrelated subspaces, the $k_t$ dimension of the state update is approximately orthogonal to the $k_{t-k}$ dimensions carried in $S_{t-k}$, and $g_{\text{past}}$ would be structurally orthogonal to $g_{\text{imm}}$ through this subspace separation. The hypothesis predicts $|\cos|$ grows monotonically with input autocorrelation: when past inputs resemble present inputs ($k_t \approx k_{t-k}$), the subspaces collapse and orthogonality breaks down.

We test this with an AR(1) input-correlation sweep. Inputs are $x_t = \rho x_{t-1} + \sqrt{1 - \rho^2}\, \epsilon_t$ with $\rho \in \{0, 0.3, 0.6, 0.9, 0.99\}$ and $\epsilon_t \sim \mathcal{N}(0, 1)$. Targets are a $50$-step delayed copy, 10 seeds per $\rho$ on vanilla RNN ($n{=}64$) and RetNet ($n{=}64$). We measure $\cos(g_{\text{imm}}, g_{\text{past}})$ at the end of a $200$-step TBPTT window.

\begin{table}[h]
\centering
\caption{$\cos(g_{\text{imm}}, g_{\text{past}})$ across input autocorrelation $\rho$, $10$ seeds, bootstrap $95\%$ CI. RetNet cos remains small across $\rho \in [0, 0.9]$; at $\rho{=}0.99$ the sign reverses (wrong direction under the hypothesis). The decorrelation hypothesis is falsified.}
\label{tab:v93}
\small
\begin{tabular}{lcc}
\toprule
$\rho$ & Vanilla RNN cos & RetNet cos \\
\midrule
$0.00$ & $-0.001$ $[-0.013, +0.012]$ & $-0.000$ $[-0.003, +0.003]$ \\
$0.30$ & $-0.004$ $[-0.015, +0.005]$ & $+0.002$ $[-0.009, +0.015]$ \\
$0.60$ & $-0.010$ $[-0.032, +0.005]$ & $+0.019$ $[-0.009, +0.049]$ \\
$0.90$ & $+0.021$ $[+0.012, +0.030]$ & $+0.002$ $[-0.075, +0.071]$ \\
$0.99$ & $+0.009$ $[-0.002, +0.024]$ & $-0.125$ $[-0.243, -0.008]$ \\
\bottomrule
\end{tabular}
\end{table}

Under the decorrelation hypothesis we would expect $\cos \to +1$ as $\rho \to 1$: past and present inputs become identical, subspace separation collapses, and the state accumulator's contribution aligns with the immediate derivative. We observe the opposite. RetNet $\cos$ stays small through $\rho \in [0, 0.9]$ and at $\rho = 0.99$ takes a wrong-sign value of $-0.125$. The vanilla-RNN comparison shows no systematic $\rho$-dependence either. Input decorrelation is not the mechanism behind small $|\cos|$ on non-isotropic architectures; an alternative structural explanation is open.

\section{Cross-architecture structural replication: LSTM and RetNet}
\label{app:lstm_cross_arch}

We replicated the structural measurement (Section~\ref{sec:orthogonality}) on LSTM ($H{=}64$ cell state) and RetNet ($H{=}64$) at matched BCI cross-session protocol, f64 AVX-2, 5 seeds per cell.

\paragraph{LSTM, 4 optimizers, per-optimizer LR-swept best.} Phase 1: LR sweep over $\{1{\times}10^{-4}, 3{\times}10^{-4}, 1{\times}10^{-3}, 3{\times}10^{-3}, 1{\times}10^{-2}\}$, 2 seeds. Phase 2: selected best LR per optimizer, 5 seeds. All four LSTM cells fail the pre-registered primary threshold (top-1 $\geq 0.50$ with gap $\geq 0.20$).

\begin{table}[h]
\centering
\caption{LSTM $H{=}64$ BCI cross-session, f64 AVX-2, 5 seeds per optimizer at per-optimizer LR-swept best. All four cells fail the pre-registered primary kill (top-1 $\geq 0.50$ and top-1 gap vs.\ $G_{\text{imm}} \geq 0.20$). Compare RNN equivalents of $0.62$--$0.74$ and $+0.41$. The rank-1 signature is specific to vanilla RNN's additive linear recurrence.}
\label{tab:lstm_cross_arch}
\footnotesize
\begin{tabular}{lccccc}
\toprule
Opt. & LR & Recovery (\%, CI) & top1$(G_{\text{past}})$ & top1$(G_{\text{imm}})$ & gap \\
\midrule
SGD       & $3{\times}10^{-4}$ & $+45.7\;[+41.2, +50.9]$ & $0.361$ & $0.348$ & $+0.013$ \\
Adam      & $3{\times}10^{-3}$ & $+63.8\;[+62.7, +64.9]$ & $0.299$ & $0.264$ & $+0.035$ \\
LARS      & $1{\times}10^{-2}$ & $+52.9\;[+48.8, +56.0]$ & $0.403$ & $0.345$ & $+0.058$ \\
Adafactor & $3{\times}10^{-3}$ & $+66.4\;[+63.3, +69.0]$ & $0.364$ & $0.312$ & $+0.052$ \\
\bottomrule
\end{tabular}
\end{table}

Three takeaways. (a) The structural claim is \emph{not} a measurement artifact on LSTM: recovery is clean ($46$--$66\%$) and LR sweeps found interior optima on Adam and Adafactor. (b) The collapse is uniform across optimizers: a $6$-- to $40$-fold drop in the top-1 gap vs RNN at matched conditions. (c) The dominant-parameter measurement is uninformative on LSTM, because the lstm.W block is $0.98$ of total parameters by count, so any direction has $\approx 0.98$ mass in lstm.W; we omit that column.

LSTM+Adam is the worst-structural/best-behavioral pairing: highest recovery ($+63.8\%$) and lowest top-1 concentration ($0.299$) of the four optimizers. That is consistent with the mechanism of Section~\ref{sec:orthogonality}: if $g_{\text{past}}$ has no rank-1 direction for Adam's per-coordinate normalization to differentially flatten, Adam has no disadvantage vs full RTRL, and recovery is set by overall gradient quality rather than structural differentiation.

\paragraph{RetNet, RNN-tuned SGD lr.} RetNet at matched RNN-tuned SGD lr$=10^{-4}$ diverges: adaptation makes held-out MSE $\sim 2\times$ worse than frozen (recovery $-210\%$). The forward-pass condition number of the parameter Jacobian is $\sim 10^4$ for RetNet (Figure~\ref{fig:grad_ratio}), three orders of magnitude above vanilla RNN's $[2, 7]$ range; an LR tuned on low-conditioning RNN dynamics is mis-scaled for RetNet's multiplicative key-value structure. The SVD of the divergent trajectory's $g_{\text{past}}$ concentrates at top-1 $= 0.77$ with $0.85$ mass in $W_{in}$, but that is the blow-up direction, not a structural property of a converged run; we do not interpret structural measurements on divergent trajectories. RetNet's structural scope is untested under the RNN-calibrated protocol, and per-architecture LR calibration (Section~\ref{sec:orthogonality}) is a prerequisite we leave to future work.

\section{Capacity-matched LSTM cross-check}
\label{app:v103d2_capacity_match}

LSTM $H{=}32$'s four-gate $h{\to}h$ matrix has $4 \cdot 32^2 = 4096$ parameters, matching RNN $H{=}64$'s $W_{hh}$. Setup: BCI cross-session, 5 seeds, f64 AVX-2, $80/20$ split, signSGD, LR grid $\{10^{-5},\ldots,10^{-2}\}$. fr-best LR is interior at $3{\times}10^{-3}$ (the $10^{-2}$ extension gives $+48.11$\,pp, below $+57.45$). $d{=}0$-best is $10^{-4}$.

\begin{table}[h]
\centering
\caption{signSGD fr-vs-$d{=}0$ gap at matched hidden-to-hidden recurrent capacity. LSTM $H{=}32$ has $4 \cdot 32^2 = 4096$ h$\to$h parameters, identical to RNN $H{=}64$'s $W_{hh}$. The gap at LSTM $H{=}32$ is within CI of the gap at LSTM $H{=}64$, and both are $\approx 2\times$ RNN. $5$ seeds, $95\%$ bootstrap CIs, BCI f64, per-method fr/$d{=}0$ best LR.}
\label{tab:v103d2_capacity}
\footnotesize
\begin{tabular}{lcccccc}
\toprule
config & h$\to$h params & fr LR & fr rec & $d{=}0$ rec & gap (pp, 95\% CI) & abs MSE benefit \\
\midrule
RNN $H{=}64$  & 4096  & $3{\times}10^{-3}$ & $+35.59$ & $+17.83$ & $+17.76$ [$+14.00, +21.47$] & $+0.119$ \\
LSTM $H{=}32$ & 4096  & $3{\times}10^{-3}$ & $+57.45$ & $+23.05$ & $+34.40$ [$+32.47, +36.37$] & $+0.255$ \\
LSTM $H{=}64$ & 16384 & $3{\times}10^{-3}$ & $+65.31$ & $+28.52$ & $+36.79$ [$+34.09, +39.15$] & $+0.286$ \\
\bottomrule
\end{tabular}
\end{table}

At matched $h{\to}h$ capacity, LSTM's gap is $\approx 2\times$ RNN's in both recovery percent (ratio $1.94$) and absolute MSE benefit (ratio $2.15$). Drift baselines differ by only $1.16\times$, so the ratio is not a denominator artifact. Doubling capacity to LSTM $H{=}64$ changes the gap from $+34.40$ to $+36.79$\,pp (within CI); parameter count is ruled out.

\section{Substrate-invariance control for magnitude dominance}
\label{app:v103_substrate}

Is the magnitude dominance specific to signSGD's trajectory, or substrate? We re-measured under plain SGD at its fr-best LR ($10^{-4}$), same pretrained $\theta$ and adapt sequence, 5 seeds, f64, $41$ TBPTT windows.

The cross-architecture ratio (RNN)/(LSTM) matches to two decimal places under both: $4.19$ (signSGD) vs $4.20$ (SGD). Absolute magnitudes differ by $\sim 1.4\times$ between optimizers on each architecture (Table~\ref{tab:v103_substrate}), consistent with signSGD's larger effective steps producing more variable per-window $g_{\text{past}}$. The architectural \emph{difference}---RNN has $\approx 4\times$ LSTM's dominance---is preserved under both, and we treat it as substrate-invariant.

\begin{table}[h]
\centering
\caption{Per-coordinate structural measurements on recurrent parameters (RNN $W_{hh}$ and LSTM $H{=}64$ \texttt{lstm.W}), BCI cross-session, f64, $41$ TBPTT windows at $W{=}200$, $5$ seeds, $95\%$ bootstrap CIs. Columns under signSGD and under SGD use each optimizer's fr-best LR. Cross-architecture ratio of $|g_{\text{past}}|/|g_{\text{imm}}|$ preserved to two decimal places across optimizers ($4.19$ vs $4.20$).}
\label{tab:v103_substrate}
\footnotesize
\begin{tabular}{lcccc}
\toprule
metric & RNN / signSGD & RNN / SGD & LSTM / signSGD & LSTM / SGD \\
\midrule
$|g_{\text{past}}|/|g_{\text{imm}}|$ median &
$20.6$ [$11.5, 31.5$] & $14.4$ [$6.1, 22.8$] & $4.93$ [$4.59, 5.28$] & $3.44$ [$3.18, 3.64$] \\
past-wins-magnitude fraction &
$1.000$ & $0.981$ & $0.940$ & $0.885$ \\
$\sigma(\mathrm{sign}\,g_{\text{past}})$ &
$0.134$ & $0.258$ & $0.152$ & $0.334$ \\
$\mathrm{sign}(g_{\text{tot}}) \ne \mathrm{sign}(g_{\text{imm}})$ rate &
$0.376$ & $0.351$ & $0.326$ & $0.290$ \\
\midrule
\multicolumn{5}{l}{\textbf{Cross-arch ratio (RNN)/(LSTM): $4.19$ (signSGD) vs $4.20$ (SGD) --- matches to two decimal places}} \\
\bottomrule
\end{tabular}
\end{table}

\section{LoRA adaptation details (preliminary; adapt-window only)}
\label{app:lora}

\paragraph{Caveat.} All recoveries in this appendix are measured on the adaptation stream (adapt-window). No held-out evaluation was run on LoRA. Because the BCI analysis in this paper (Appendix~\ref{app:heldout}) establishes that adapt-stream recovery can reflect overfitting to the adaptation stream rather than genuine generalization (Adafactor, SGD, and LARS all show large adapt-stream gains that collapse or invert on held-out data), the numbers below are reported as preliminary scaling evidence and should not be read as method or optimizer comparisons. Held-out LoRA validation is future work.

\begin{table}[h]
\centering
\caption{Online LoRA adaptation across pretrained LMs with $d{=}0$ + Adam at lr$=10^{-3}$ (rank-4 adapters, code-to-Wikipedia shift, single-pass over 20K tokens with shift at 10K, no replay). \textbf{Adapt-window recovery only; no held-out validation was run. Subject to the stream-fitting confound documented in Appendix~\ref{app:heldout}; see caveat above.} 5 seeds for Qwen2, 3 seeds for others (single-seed reported as point estimate). Baseline denominator is full RTRL+Adam on the same model with truncated-BPTT ($w{=}64$) as a practical proxy; Mamba-1.4B's $133.5\%$ therefore reflects exceeding this proxy, not an exact RTRL baseline (full closed-form RTRL on a 1.4B-parameter model is computationally infeasible and never run).}
\label{tab:lora}
\begin{tabular}{lcc}
\toprule
Model & Params & Recovery ($d{=}0$ + Adam) \\
\midrule
Mamba-1.4B & 1.4B & $\mathbf{133.5\%}$ \\
Mamba-130M & 130M & $51.6\%$ \\
TinyLlama & 1.1B & $31.6\%$ \\
Qwen2 & 7B & $31.3 \pm 7.3\%$ \\
GPT-2 & 124M & $15.0 \pm 0.9\%$ \\
\bottomrule
\end{tabular}
\end{table}

\paragraph{Optimizer choice on LoRA.} On GPT-2 (the model where Adam-only recovery is weakest), we ran a per-optimizer LR sweep to test whether the BCI optimizer pattern (Adafactor outperforms Adam on real drift) appears here too. It does: Adafactor at lr$=3{\times}10^{-3}$ achieves $26.5 \pm 4.3\%$ recovery vs Adam's $15.0 \pm 0.9\%$ at the same setup (3 seeds, best of $\{3{\times}10^{-4}, 10^{-3}, 3{\times}10^{-3}\}$ for Adafactor; absolute post-shift cross-entropy $3.65$ vs $3.77$, against frozen baseline $3.94$ and pre-shift $2.83$). The BCI ``Adam is suboptimal'' pattern extends to small transformer LoRA. We did not test alternative optimizers on the larger Mamba/TinyLlama/Qwen2 models.

We tested SGD with the same lr$=10^{-3}$ as a baseline; recovery was near-zero across all five models. We do not interpret this as evidence that SGD is mechanistically incapable of adapting LoRA adapters: per-optimizer LR tuning was not performed at the LoRA scale (results on smaller recurrent networks in Appendix~\ref{app:optimizers} show that SGD requires substantially higher learning rates than Adam to adapt). Whether SGD with appropriate per-task LR tuning matches Adam on LoRA adaptation is an open question we leave to future work.

With rank-16 adapters and 40K tokens, GPT-2 recovery reaches 136.8\% (cross-entropy 2.744, down from 3.493 frozen). LoRA gradient norms on Mamba across 24 layers and 3 matrix types show a $51.8\times$ magnitude ratio between the largest group (L23 dt\_proj, norm 1.35) and smallest (L15 x\_proj, norm 0.026). Last-layer gradients are $12\times$ larger than mid-network.

\paragraph{Multi-domain sequential shifts.} Streaming Code$\to$Wiki$\to$Code$\to$Wiki through GPT-2 with a rank-4 adapter and Adam, the adapter adapts to every shift (39\% and 26\% recovery on successive Wiki segments). The adapter does not forget: Code B cross-entropy (1.608) is lower than Code A (1.796), indicating improvement on return to a previously seen domain. Wiki readaptation is faster on second exposure (shock cross-entropy 2.612 vs 3.048). No replay buffer, no task boundaries, single-pass.

\section{Streaming ML benchmarks}
\label{app:streaming}

Online RNNs with decay 0.0 and Adam outperform tree-based streaming methods on tasks dominated by continuous drift. On the Hyperplane generator from the River library \citep{montiel2021river} (200K samples, 10 features, gradual drift, parameters matching \citealt{gomes2017arf}), the RNN achieves 0.925 accuracy versus Adaptive Random Forest at 0.842 ($+8.3$ pp). Tree ensembles retain their advantage on tasks requiring spatial partitioning (RandomRBF) and real-world tabular data (Electricity).

\section{Test-time adaptation in vision}
\label{app:tta}

On CIFAR-10-C test-time adaptation (WideResNet-28-10, 15 corruption types at severity 5, 3 seeds), adapting only batch normalization affine parameters via entropy minimization \citep{wang2021tent} shows a 2.6 percentage point gap between Adam and SGD with per-optimizer LR tuning. BN parameters are the same type at every layer, with roughly uniform gradient scales. Replacing BN with LoRA adapters at every convolutional layer widens the Adam--SGD gap to 7.6 pp (51.1\% vs 43.5\% recovery), with LoRA gradients spanning 28 layers and a 5--6$\times$ norm ratio between shallowest and deepest. $\beta_2$-only matches Adam (52.3\%); $\beta_1$-only falls between (46.0\%). This is the only test-time adaptation setting in our experiments where Adam meaningfully outperforms tuned SGD; on the recurrent online-adaptation tasks studied in the main paper, multiple optimizers are competitive (Appendix~\ref{app:optimizers}).

\section{Long-horizon adaptation}
\label{app:horizon}

To test whether the 50-step dependency limit in our earlier experiments was a method limit or an architecture limit, we ran the delayed sine task at $t{+}50$, $t{+}100$, $t{+}200$, and $t{+}500$ on both vanilla RNN ($n{=}64$) and LSTM ($n{=}64$), with per-method LR tuning at each horizon. $d{=}0$ + Adam matches or exceeds full RTRL at every delay.

\begin{table}[h]
\centering
\caption{Long-horizon delayed sine prediction. Post-shift MSE (5 seeds). $d{=}0$ + Adam matches or beats RTRL at every delay, up to $t{+}500$.}
\label{tab:horizon}
\small
\begin{tabular}{lccccc}
\toprule
Architecture & Delay & $d{=}0$ + Adam & Full RTRL & Frozen & Ratio \\
\midrule
Vanilla RNN & $t{+}50$ & 0.0044 & 0.0031 & 1.25 & $\approx$ \\
Vanilla RNN & $t{+}100$ & 0.31 & 0.32 & 0.71 & $\approx$ \\
Vanilla RNN & $t{+}200$ & \textbf{0.065} & 0.083 & 0.69 & $0.8\times$ \\
Vanilla RNN & $t{+}500$ & 0.0079 & 0.0077 & 0.31 & $\approx$ \\
\addlinespace
LSTM & $t{+}50$ & \textbf{0.0052} & 0.0082 & 0.65 & $0.6\times$ \\
LSTM & $t{+}100$ & \textbf{0.19} & NaN (diverged) & 1.72 & -- \\
LSTM & $t{+}200$ & \textbf{0.0068} & 0.020 & 0.71 & $0.3\times$ \\
LSTM & $t{+}500$ & \textbf{0.0042} & 0.017 & 0.18 & $0.2\times$ \\
\bottomrule
\end{tabular}
\end{table}

On vanilla RNN, $d{=}0$ and RTRL are statistically indistinguishable at all horizons, with $d{=}0$ slightly better at $t{+}200$. On LSTM, $d{=}0$ beats RTRL at every horizon by factors of $0.2$--$0.6\times$; at $t{+}100$, RTRL diverges (NaN) while $d{=}0$ stays stable. That stability advantage is a practical benefit beyond the $1000\times$ memory savings: full RTRL's $O(n^4)$ sensitivity tensor is both expensive and numerically fragile on gated architectures. Pre-shift MSE is high at longer delays ($0.5$--$0.65$) because the architecture cannot fully memorize a $t{+}500$ sinusoid, but adaptation still works: at $t{+}500$, $d{=}0$ reaches post-shift MSE $0.008$ vs frozen $0.18$--$0.31$, a $20$--$40\times$ reduction.

\section{RWKV boundary condition}
\label{app:rwkv}

On chaotic dynamics, $\beta_2$-containing optimizers (Adam $\beta_2$-only, full Adam, RMSprop) diverge on every seed on RWKV, while SGD is the only survivor (293\% recovery, zero variance). Under SGD, RWKV's recurrent gradients are near-zero (norm $\sim$$10^{-4}$ at initialization, $\sim$$10^{-6}$ after training, vs output norm $\sim$$10^{-1}$), so SGD effectively freezes the recurrence and adapts only through the output predictor. Direct measurement of Adam's internal state confirms the failure mechanism is gradient scale \emph{amplification}: $\beta_2$ normalization amplifies recurrent gradients by $2 \times 10^5$-fold per step (raw $\|g_\mathrm{rec}\| = 5 \times 10^{-3}$, effective $\|g_\mathrm{rec} / \sqrt{\hat{v}_t}\| = 686$), flipping the gradient-scale hierarchy from recurrent-to-output ratio $0.008$ (raw) to $13.1$ (effective). That creates a positive feedback loop: amplified recurrent updates increase recurrent gradient magnitude, which further amplifies subsequent updates, causing divergence within 17~steps on every seed. This is distinct from the compression that $\beta_2$ normally corrects; here normalization destabilizes because the multiplicative $k \cdot v$ state update amplifies the overscaled recurrent updates on chaotic dynamics.

\section{Truncated BPTT comparison}
\label{app:tbptt}

A natural question is whether truncated BPTT with a longer window outperforms immediate derivatives ($w{=}1$). We compared $w \in \{1, 4, 16, 64\}$ on sine, Lorenz, and cross-session BCI, all starting from the same converged model (trained with $w{=}1$ pre-shift, window varied only at adaptation time). Table~\ref{tab:tbptt} reports post-shift MSE. All short windows ($w{=}1$ through $w{=}4$) adapt successfully; longer windows degrade or fail.

\begin{table}[h]
\centering
\caption{Truncated BPTT window comparison (5 seeds, converged pre-shift model, window varied at shift only). Values are post-shift MSE; ratio to $w{=}1$ in parentheses. Lower is better.}
\label{tab:tbptt}
\small
\begin{tabular}{lcccc}
\toprule
Task & $w{=}1$ & $w{=}4$ & $w{=}16$ & $w{=}64$ \\
\midrule
Sine & 0.0009 (1.0$\times$) & \textbf{0.0005} (0.5$\times$) & 0.0013 (1.5$\times$) & 0.0024 (2.7$\times$) \\
Lorenz & 0.00009 (1.0$\times$) & \textbf{0.00003} (0.3$\times$) & 0.00006 (0.6$\times$) & 0.001 (12$\times$) \\
BCI (Adam) & 0.339 (1.0$\times$) & 0.321 (0.9$\times$) & \textbf{0.294} (0.9$\times$) & 0.356 (1.0$\times$) \\
BCI (RMSprop) & \textbf{0.254} (1.0$\times$) & 0.242 (1.0$\times$) & 0.275 (1.1$\times$) & 0.309 (1.2$\times$) \\
\bottomrule
\end{tabular}
\end{table}

Both $w{=}1$ and $w{=}4$ adapt on every task: on Lorenz, both recover $>$99\% relative to frozen (MSE 0.00009 and 0.00003 respectively, vs frozen MSE 0.051); on sine, both reduce post-shift MSE below 0.001 (vs frozen 0.004). The differences between $w{=}1$ and $w{=}4$ are small relative to the frozen--adapted gap. Longer windows degrade: $w{=}16$ is worse than $w{=}1$ on sine and BCI-RMSprop, and $w{=}64$ is universally harmful (12$\times$ worse on Lorenz, diverges on sine when training online). In fully online training (from scratch with each window size), $w{\geq}4$ also suffers from fewer parameter updates per timestep, and $w{=}64$ diverges on sine (5/5 seeds). This fits the paper's main finding: temporal credit beyond a few steps adds noise, not signal, and the expensive Jacobian propagation of full RTRL is unnecessary.


\section*{NeurIPS Paper Checklist}

\begin{enumerate}

\item {\bf Claims}
    \item[] Answer: \answerYes{}
    \item[] Justification: The abstract scopes the sufficiency claim to specified cells (vanilla-RNN synthetics under Adam and SGD, LSTM/sine/Adam, and BCI f64 $50/50$ with Adam), and explicitly names the four exceptions (LARS fr-advantage as optimizer$\times$method interaction, SGD $80/20$ split-dependence, Adafactor hardware-inconsistency, float32 kernel-dispatch fragility). All claims are tested with 5 seeds; the BCI $50/50$ equivalence uses $n{=}20$ with a pre-registered TOST at $\pm 3$\,pp. Contributions with Does-claim / Does-not-claim scope are listed explicitly in Section~1.

\item {\bf Limitations}
    \item[] Answer: \answerYes{}
    \item[] Justification: Section~\ref{sec:discussion} covers: the precision- and LR-tuning-dependence of the BCI comparison (float32 full-RTRL recovery depends on CPU kernel dispatch; at f64 with true per-hardware best LR for both methods, Adam is equivalent, SGD $d{=}0$-wins by $7$--$11$\,pp, LARS is full-RTRL-better by $+16$ to $+25$\,pp with $d{=}0 +$ LARS independently failing, Adafactor hardware-inconsistent with a $\approx 8$\,pp cross-hardware sign-flip); optimizer-specific LR portability (Adam hardware-portable, Adafactor/LARS shift $3\times$); an initial-grid-coarseness artifact on $d{=}0$ Adafactor and $d{=}0$ LARS at f64, resolved with finer sweeps; the open structural mechanism for small per-step $|\cos|$ on non-isotropic architectures (isotropy and input-decorrelation hypotheses falsified); the failed per-step direction-causal intervention at $n{=}10$ (Spearman $+0.297$ vs pre-reg $+0.50$); and pretrained-model scope (LoRA only, no held-out validation at LoRA scale). Three pre-registered tests of candidate mechanism extensions (preconditioner-alignment, optimizer-class taxonomy, LSTM sign-based test) all falsified their specific hypotheses; the structural mechanism in Section~\ref{sec:orthogonality} is supported as a measured correspondence rather than a derived prediction.

\item {\bf Theory assumptions and proofs}
    \item[] Answer: \answerNA{}
    \item[] Justification: The paper is empirical. The small-$|\cos|$ observation is documented as a phenomenon; two candidate structural explanations (isotropy, input decorrelation) were tested and falsified (Section~\ref{sec:orthogonality}), and no formal proof is offered.

\item {\bf Experimental result reproducibility}
    \item[] Answer: \answerYes{}
    \item[] Justification: All architectures, hyperparameters, seeds, and task specifications are stated in each section. Code will be released upon acceptance.

\item {\bf Open access to data and code}
    \item[] Answer: \answerYes{}
    \item[] Justification: Code will be released upon acceptance. Synthetic tasks are fully specified. BCI data is publicly available (Zenodo DOI: 10.5281/zenodo.3854034). Streaming ML uses the River library.

\item {\bf Experimental setting/details}
    \item[] Answer: \answerYes{}
    \item[] Justification: Hidden sizes, learning rates, optimizer configurations, number of seeds, and task parameters are specified throughout. Full details in Appendix.

\item {\bf Experiment statistical significance}
    \item[] Answer: \answerYes{}
    \item[] Justification: Main results report mean with $95\%$ bootstrap or t-based CI across 5 seeds; the Adam $50/50$ BCI equivalence claim uses $n{=}20$ seeds and a pre-registered TOST at $\pm 3$\,pp margin (Section~\ref{sec:scaling}). Error bars in figures capture seed variability.

\item {\bf Experiments compute resources}
    \item[] Answer: \answerYes{}
    \item[] Justification: All from-scratch experiments run on a single NVIDIA T4 GPU (16\,GB). LoRA experiments on a single T4 (Kaggle). Total compute is approximately 50 GPU-hours including preliminary experiments.

\item {\bf Code of ethics}
    \item[] Answer: \answerYes{}
    \item[] Justification: The research conforms with the NeurIPS Code of Ethics. No human subjects, no personally identifiable data.

\item {\bf Broader impacts}
    \item[] Answer: \answerNA{}
    \item[] Justification: This is foundational research on optimization for recurrent networks. There is no direct path to negative societal applications beyond those inherent to improving neural network training.

\item {\bf Safeguards}
    \item[] Answer: \answerNA{}
    \item[] Justification: The paper does not release pretrained models or scraped datasets. All models are small ($\leq$7B) and trained on synthetic or public data.

\item {\bf Licenses for existing assets}
    \item[] Answer: \answerYes{}
    \item[] Justification: BCI data is from Zenodo (CC-BY). River library is BSD-3. Pretrained models (GPT-2, TinyLlama, Mamba, Qwen2) are used under their respective open licenses.

\item {\bf New assets}
    \item[] Answer: \answerNA{}
    \item[] Justification: The paper does not release new datasets or pretrained models. Code will be released upon acceptance.

\item {\bf Crowdsourcing and research with human subjects}
    \item[] Answer: \answerNA{}
    \item[] Justification: No crowdsourcing or human subjects research.

\item {\bf Institutional review board (IRB) approvals or equivalent for research with human subjects}
    \item[] Answer: \answerNA{}
    \item[] Justification: No human subjects research.

\item {\bf Declaration of LLM usage}
    \item[] Answer: \answerNA{}
    \item[] Justification: LLMs are not used as a component of the core methods. LLM assistance was used for writing and editing only.

\end{enumerate}

\end{document}